%% file: main.tex
\documentclass{article}

\PassOptionsToPackage{numbers,compress}{natbib}
\usepackage[final]{neurips_2026}
\makeatletter
\renewcommand{\@noticestring}{Workshop on Who Verifies the Agents? Toward Reliable Agent Development, NeurIPS 2026.}
\makeatother

\usepackage[utf8]{inputenc}
\usepackage[T1]{fontenc}
\usepackage{hyperref}
\hypersetup{
  hidelinks,
  pdftitle={Agent Verifiers in Counterfactual Analysis: A Pilot Study Benchmarking Agentic Heterogeneous Treatment Effects},
  pdfauthor={Satiyabooshan Murugaboopathy, Entejar Alam, Adel Daoud, Connor T. Jerzak}
}
\usepackage{url}
\usepackage{booktabs,longtable,pdflscape,array}
\usepackage{amsfonts}
\usepackage{nicefrac}
\usepackage{microtype}
\usepackage{xcolor}
\usepackage{amsmath,amssymb,amsthm,bm}
\usepackage{mathtools}
\usepackage{graphicx}
\graphicspath{{./}}
\usepackage{caption}
\usepackage{tikz}
\usetikzlibrary{arrows.meta,positioning,calc}
\usepackage{pgfplots}
\pgfplotsset{compat=1.17}

\definecolor{ink}{HTML}{1A2733}
\definecolor{deep}{HTML}{0F4C5C}
\definecolor{accent}{HTML}{E36414}
\definecolor{fog}{HTML}{E8ECEF}
\definecolor{tabbar}{HTML}{B9C6CF}

\newtheorem{theorem}{Theorem}
\newtheorem{lemma}{Lemma}
\newtheorem{corollary}{Corollary}
\theoremstyle{definition}
\newtheorem{assumption}{Assumption}
\theoremstyle{remark}
\newtheorem{remark}{Remark}

\newcommand{\Filt}{\mathcal{F}_{i,t_i^-}}
\newcommand{\indep}{\perp\!\!\!\perp}
\newcommand{\Dpre}{\mathcal{D}_{i,t_i^-}}
\newcommand{\Cpre}{\mathcal{C}_{t_i^-}}
\newcommand{\Gpost}{\mathcal{G}_{\ge t_i}}
\newcommand{\KL}[2]{D\!\left(#1\,\middle\|\,#2\right)}
\DeclareMathOperator{\I}{I}
\newcommand{\Alg}{\mathcal{A}}

\title{Agent Verifiers in Counterfactual Analysis: \\ A Pilot Study Benchmarking Agentic Heterogeneous Treatment Effects}
\workshoptitle{Who Verifies the Agents? Toward Reliable Agent Development}

\author{%
  Satiyabooshan Murugaboopathy$^{1,2}$  \quad Entejar Alam$^{6}$ \\
  \textbf{Adel Daoud$^{2,3,4}$ \quad Connor T.\ Jerzak$^{2,5,*}$} \\[6pt]
  \small $^1$Institute for Analytical Sociology, Link\"oping University, Link\"oping, Sweden \\
  \small $^2$AI and Global Development Lab \quad $^3$Uppsala University, Uppsala, Sweden \\
  \small $^4$NLP@DSAI, Chalmers University of Technology, Gothenburg, Sweden \\
  \small $^5$Department of Government, The University of Texas at Austin, Austin, TX, USA \\
  \small $^6$Department of Statistics and Data Sciences, The University of Texas at Austin, Austin, TX, USA \\
  \small $^*$Corresponding author: \texttt{connor.jerzak@austin.utexas.edu}
}

\input{FiguresMake/generated/numbers}

\input{FiguresMake/generated/beta_numbers}
\makeatletter
\newcommand{\scinum}[1]{\edef\scinum@tmp{#1}\expandafter\@scinum\scinum@tmp e\@nil}
\def\@scinum#1e#2\@nil{\ifx\relax#2\relax$#1$\else\@scinum@exp{#1}#2\fi}
\def\@scinum@exp#1#2e{$#1\times 10^{\number\numexpr#2\relax}$}
\makeatother

\begin{document}

\maketitle

\begin{abstract}
Agents can expand the covariate space of randomized experiments by retrieving fast-moving local context missed by slow-changing survey-obtained baselines such as age and gender. Such covariates would help characterize who benefits most from an intervention. Yet verifying such covariates is difficult: individual treatment effects are unobserved, retrieval leakage can pull post-treatment information into nominally pretreatment summaries, and encoder (neural-weights-based) leakage can inject post-treatment pretraining knowledge into representations even when input text passes temporal filtering. Guided by a two-channel information bound separating summary provenance from encoder weights, we propose a three-layer benchmark protocol to verify agentic treatment-effect representations for causal analysis: (i) \emph{retrieval admissibility}, enforced by pre-treatment date filtering; (ii) \emph{heterogeneity validation}, scored out of sample by rank-weighted average treatment effect (RATE) ratios; and (iii) \emph{encoder admissibility}, verified with chronologically vintaged encoders, future-fact probes, explicit-label controls, and weight perturbations. Across \BatteryStudyCount{} datasets (\BatteryContrastCount{} treatment--outcome contrasts) anchored by Uganda's Youth Opportunities Program, we compare \emph{session-isolated retrieval}---sandboxed, bounded-budget agent runs per study context---against a \emph{persistent workspace teammate} (xAI's GrokBot) that builds one date-verified corpus per study while blinded to outcomes. The workspace teammate yields higher mean RATE ratios in \GrokTextHigherStudies{} of \BatteryStudyCount{} datasets for text alone and \GrokCombinedHigherStudies{} for combined features, though paired tests show neither retrieval workflow universally dominates. Even on fixed retrieved text, varying the pretrained text encoder significantly shifts downstream heterogeneity signal. Auditing this encoder channel, chronologically vintaged probes detect no post-event shift on clean historical text, yet explicit-label controls expose sharp sensitivity to token pooling and controlled single-record adapter perturbations show that memorized future facts persist until injected noise degrades model fluency. Temporal filtering at retrieval is thus necessary in principle but insufficient alone: verifying agent-generated covariates requires jointly auditing source provenance, out-of-sample treatment-effect ranking, and encoder weights.
\end{abstract}

\section{Introduction}

Randomized controlled trials (RCTs) estimate average treatment effects, but practical decisions---such as targeted interventions and personalized treatment allocation---depend on identifying who benefits most. The individual effect $\tau_i = Y_i(1)-Y_i(0)$ is not jointly observable; for baseline covariates $X_i \in \mathcal{X}$, the prediction target in conditional average treatment effect (CATE) estimation is instead $\tau_{\mathcal{X}}(x)=\mathbb{E}[\tau_i\mid X_i=x]$ \citep{athey2016recursive,wager2018estimation,kunzel2019metalearners,chernozhukov2018double}. The quality of any CATE estimate is bounded by the quality of the pretreatment covariates it conditions on. Yet rarely do researchers know in advance which signals matter most.

Recent work has expanded the covariate space of experiments with new data modalities: satellite imagery \citep{jerzak2023image,jerzak2024effect}, text \citep{egami2022make}, and vision-language representations \citep{murugaboopathy2027platonicrepresentationspovertymapping}. In particular, archived web text records fast-moving local events---political unrest, price shocks, infrastructure projects, the local context around an RCT---that slow-changing survey-obtained baseline covariates may struggle to capture. Language-model agents are a potentially promising tool for retrieving such information at large scale and low cost.

Deploying such agents in causal studies, however, poses a verification problem absent from existing agent benchmarks, which evaluate task completion, instruction following, and tool use in interactive environments \citep{liu2024agentbench} or aggregate capability metrics across tasks \citep{liang2023helm}. Rather than scoring task execution, we ask whether an agent's data-producing behavior yields representations that are causally admissible for downstream treatment-effect estimation. Checking whether agent-retrieved context improves CATE estimation is difficult because unit-level effects $\tau_i = Y_i(1)-Y_i(0)$ are unobserved and retrieved documents, summaries, or embeddings can leak post-treatment outcomes into pretreatment covariates.

Moreover, post-treatment leakage can corrupt causal inference through two distinct channels. First, under \emph{retrieval leakage}, an agent pulls post-treatment or retrospectively edited pages into nominally pretreatment summaries. Second, under neural-weight-based (encoder) leakage, a pretrained encoder injects knowledge from post-treatment pretraining corpora into embeddings even when the input text predates the intervention.

\begin{figure}[t]
\centering
\includegraphics[width=0.95\textwidth]{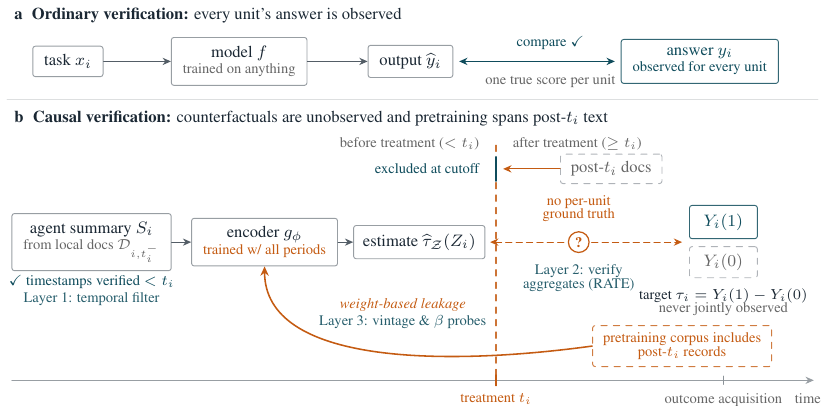}
\caption{\textbf{Verifying agentic workflows in causal systems.} \textbf{a}, In standard task benchmarks, every unit's ground-truth answer $y_i$ is observed directly. \textbf{b}, In counterfactual analysis, individual treatment effects $\tau_i = Y_i(1)-Y_i(0)$ are never jointly observed and post-$t_i$ information can leak through retrieved documents or pretrained encoder weights, motivating temporal filtering (Layer~1), aggregate RATE validation (Layer~2), and vintage and $\beta$-stability probes (Layer~3).}
\label{fig:toc}
\end{figure}

To address leakage and evaluation jointly (Fig.~\ref{fig:toc})---guided by a two-channel information bound separating summary provenance from pretraining weights (Theorem~\ref{thm:main})---we here propose a three-layer benchmark framework for \emph{agentic heterogeneous treatment effects}: \textbf{Layer~1 (retrieval admissibility)} counters retrieval leakage by restricting acquisition to verifiably pre-intervention sources; \textbf{Layer~2 (heterogeneity validation)} addresses unobserved unit-level effects and search overfitting by scoring out-of-sample prioritization with cross-fitted rank-weighted average treatment effect (RATE) ratios \citep{yadlowsky2025evaluating}; and \textbf{Layer~3 (encoder admissibility)} targets weight-based leakage by auditing pretrained encoders with chronologically vintaged models \citep{he2025chronologically}, future-fact probes, explicit-label controls, and weight perturbations.

To test how distinct agent designs perform under this protocol, we anchor our evaluation on the 2008 Uganda Youth Opportunities Program (YOP) RCT \citep{blattman2014generating,blattman2020long} within an \BatteryStudyCount{}-dataset benchmark (encompassing \BatteryContrastCount{} causal contrasts) comparing two retrieval architectures. The first is \emph{session-isolated retrieval}: a sandboxed CLI agent harness that runs ephemeral, bounded-budget web and archive sessions per study context and filters out post-cutoff dates ex post. The second is a persistent \emph{workspace teammate} (xAI's GrokBot): working in a cloud VM with shell and PDF-parsing tools, it receives study contexts without outcome data, compiles one shared corpus per study, and passes an automated date-and-keyword check before estimation begins.

In Uganda, both agentic text workflows exceed matched tabular RATE ratios on all four skill and human-capital outcomes, whereas all representations yield negative ratios on income. Across all \WorkflowComparisonN{} paired RATE tests (spanning \BatteryContrastCount{} contrasts over context-level, area-pooled, text-only, and combined specifications), only \WorkflowSignificantN{} workflow differences survive multiplicity adjustment: \WorkflowPositiveN{} favor the workspace teammate and \WorkflowNegativeN{} favor session-isolated retrieval. Meanwhile, swapping pretrained text encoders significantly shifts CATE rankings in both directions, and while chronologically vintaged probes detect no post-event shift on clean historical text, controlled weight perturbations show that memorized future facts remain recoverable until injected noise degrades model fluency.

\section{Background and problem setup}
\label{sec:setup}

\paragraph{Admissible agentic representations.} For unit $i$, let $t_i$ denote the intervention date and $X_i \in \mathcal{X}$ the raw covariates; the CATE over $\mathcal{X}$ is $\tau_{\mathcal{X}}(x) = \mathbb{E}[\tau_i \mid X_i = x]$. A representation is \emph{admissible} only if it can be constructed exclusively from information available before $t_i$; we write $\Filt$ for this verifiably pre-$t_i$ information set. Let $\mathcal{L}_i$ denote the study-defined locality of unit $i$. For each document $d$, let $\operatorname{loc}(d)$ denote its locality and $\operatorname{pubtime}(d)$ its publication or archival timestamp. The eligible historical corpus is
\begin{equation}
\Dpre
=
\big\{\, d : \operatorname{loc}(d) \in \mathcal{L}_i,\
\operatorname{pubtime}(d) < t_i \,\big\}.
\end{equation}
Let $\Theta_{\pi_i}$ collect every parameter or model weight used by the retrieval-and-summarization policy. The policy is non-anticipating only when $(\Dpre,\Theta_{\pi_i})$ is $\Filt$-measurable and
\begin{equation}
S_i = \pi_i\!\left(\Dpre,\,\Theta_{\pi_i},\, \epsilon_i\right),
\qquad
\epsilon_i \indep \left(Y_i(0),\,Y_i(1),\,X_i,\,t_i,\,
\Dpre,\,\Theta_{\pi_i}\right).
\label{eq:policy}
\end{equation}
Admissibility also requires that the policy parameters $\Theta_{\pi_i}$ do not depend on post-$t_i$ information. In practice, units in the same locality and period share a context $(\mathcal{L}_i,t_i)$. \emph{Session-isolated retrieval} applies Equation~\eqref{eq:policy} independently to each context in an ephemeral session with separate search noise $\epsilon_i$. A \emph{workspace teammate} instead pools pre-$t_i$ documents across all contexts in a study ($\bigcup_j \mathcal{D}_{j,t_j^-}$) inside one persistent workspace before mapping verified excerpts to units. Sharing one corpus across contexts reduces per-unit search noise, but any post-treatment leak in a shared source then contaminates every unit mapped to it---whereas post-$t_i$ leakage in shared policy weights ($\Theta_{\pi_i}=\Theta_{\pi}$) affects all units under either topology.

\begin{figure}[t]
\centering
\resizebox{0.94\textwidth}{!}{\input{FiguresMake/fig01_framework}}
\caption{\textbf{Three paths from experimental units to CATE representations.} \textbf{a}, Baseline representations (census covariates, Earth-observation imagery) are admissible but miss fast-changing political and economic context. \textbf{b}, Naive web retrieval captures that context but leaks future outcomes into nominally pretreatment representations. \textbf{c}, Ordered along data flow, the agentic framework filters temporal provenance at retrieval (Layer~1), audits the encoder $g_{\phi}$ for weight-based leakage (Layer~3), and scores out-of-sample heterogeneity via the RATE protocol (Layer~2).}
\label{fig:framework}
\end{figure}
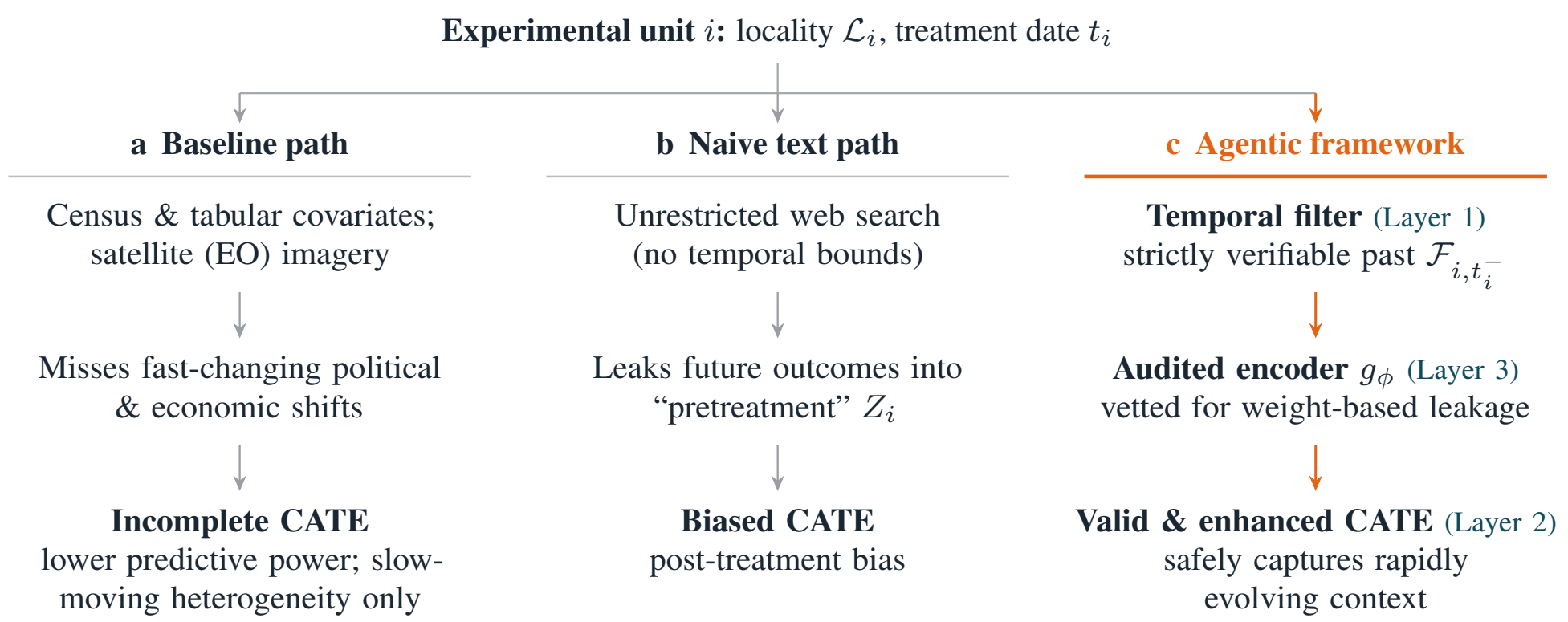

\paragraph{Encoded representations and two-channel admissibility.} Once a summary $S_i$ is assembled, mapping it through a numerical text encoder $g_{\phi}$ yields a learned representation $Z_i \in \mathcal{Z}$ and defines a CATE over $\mathcal{Z}$,
\begin{equation}
Z_i = g_{\phi}(S_i),
\qquad
\tau_{\mathcal{Z}}(z) = \mathbb{E}[\tau_i \mid Z_i = z].
\end{equation}
Because post-treatment information can enter at either stage---during text acquisition ($S_i$) or during embedding ($g_{\phi}$)---admissibility has two parts (Fig.~\ref{fig:framework}): \emph{retrieval admissibility} excludes documents or policy inputs with unverifiable timing from $S_i$, whereas \emph{encoder admissibility} requires that $g_{\phi}$ inject no knowledge learned after $t_i$ (weight-based leakage).

\paragraph{Autonomy tiers in agentic heterogeneity analysis.} Retrieving and encoding external context is one instance of a broader design space in which an agent selects covariates, engineers features, acquires new data, or optimizes the estimation pipeline itself. Let $X_i\in\mathcal{X}=\mathbb{R}^p$. For a fixed agent specification, let $J\subseteq\{1,\ldots,p\}$ be a selected covariate subset and $h_2,h_3$ measurable feature maps. Orthogonal to our three verification layers (which audit representations), we distinguish five tiers of agent autonomy (which classify what the agent constructs), yielding representations $Z_i^{(\ell)} \in \mathcal{Z}^{(\ell)}$:
\begin{equation}
Z_i^{(\ell)}=
\begin{cases}
X_i, & \ell=0 \quad\text{(researcher baseline)},\\
X_{i,J}, & \ell=1 \quad\text{(agent selection)},\\
h_2(X_i), & \ell=2 \quad\text{(agent construction)},\\
h_3(X_i,Z_i), & \ell=3 \quad\text{(agent acquisition and integration)},\\
Z_i(\widehat{\eta}), & \ell=4 \quad\text{(agent pipeline optimization)},
\end{cases}
\label{eq:agentic-levels}
\end{equation}
where $S_i$, $Z_i=g_{\phi}(S_i)$, and the historical corpus $\Dpre$ retain their definitions above, and $Z_i(\widehat{\eta})$ is the representation induced by a complete pipeline $\widehat{\eta}$ selected using performance feedback on development data. For fixed construction rules ($\ell \le 3$, or conditioning also on the development sample for $\ell=4$), each tier induces a CATE,
\begin{equation}
\tau_{\mathcal{Z}^{(\ell)}}(z)
=\mathbb{E}\!\left[\tau_i\mid Z_i^{(\ell)}=z\right].
\label{eq:agentic-cate}
\end{equation}

Moving up these autonomy tiers progressively expands the agent's scope. Tiers~1 and~2 select or nonlinearly transform existing survey variables ($X_{i,J}$ and $h_2(X_i)$), whereas Tier~3 adds historical context $Z_i$ from outside the recorded covariates: session-isolated retrieval and GrokBot both operate at Tier~3, differing in session persistence, tooling, and date verification, and are evaluated below via standalone text $h_3(x,z)=z$ (recovering $\tau_{\mathcal{Z}}$) and combined features $h_3(x,z)=(x,z)$ (inducing $\tau_{\mathcal{X},\mathcal{Z}}$ below). Tier~4 iteratively revises these choices together with the CATE estimator (formalized as RATIO in Appendix~\ref{app:ratio}); for this learned rule, the CATE is conditional on the development data used to select and fit the pipeline. Across all five tiers, every input and model weight must remain admissible before $t_i$, and any outcome-guided search must be confined to training splits (Section~\ref{sec:protocol}).

\paragraph{Additional heterogeneity resolved by acquired covariates.} Stepping from internal transformations ($\ell \le 2$) to external acquisition ($\ell = 3$) enlarges the baseline conditioning set along a nested ordering of $\sigma$-algebras (whereas Verification Layers~1--3 below are complementary checks across distinct pipeline components): for fixed selection and construction rules, the generated $\sigma$-algebras satisfy
\[
\sigma(X_{i,J})\subseteq\sigma(X_i)\subseteq\sigma(X_i,Z_i),
\qquad
\sigma\!\left(X_i,h_2(X_i)\right)=\sigma(X_i).
\]
Thus, constructing features from $X_i$ can aid finite-sample estimation without expanding the information in the full baseline, whereas acquiring external features $Z_i$ can expand it. Assuming $\mathbb{E}[\tau_i^2]<\infty$, define $\tau_{\mathcal{X},\mathcal{Z}}(x,z)=\mathbb{E}[\tau_i\mid X_i=x,Z_i=z]$. For a fixed retrieval-and-summarization specification $\pi=(\pi_i,\Theta_{\pi_i})_i$ and encoder $g_{\phi}$ (with weights $\phi$), define the additional resolved heterogeneity as
\begin{equation}
\begin{aligned}
\Gamma_{\pi,\phi}
&=\operatorname{Var}\!\left[\tau_{\mathcal{X},\mathcal{Z}}(X_i,Z_i)\right]
 -\operatorname{Var}\!\left[\tau_{\mathcal{X}}(X_i)\right]\\
&=\mathbb{E}\!\left[
  \bigl(\tau_{\mathcal{X},\mathcal{Z}}(X_i,Z_i)
  -\tau_{\mathcal{X}}(X_i)\bigr)^{2}\right]\ge 0.
\end{aligned}
\label{eq:agentic-gain}
\end{equation}
The equality follows from $\mathbb{E}[\tau_{\mathcal{X},\mathcal{Z}}(X_i,Z_i)\mid X_i]=\tau_{\mathcal{X}}(X_i)$ and the law of total variance. Consequently, $\Gamma_{\pi,\phi}>0$ exactly when the acquired covariates resolve conditional mean treatment-effect differences beyond $X_i$. Note that $\Gamma_{\pi,\phi}>0$ characterizes incremental signal of $(X_i,Z_i)$ over $X_i$ (not standalone text performance $Z_i$). To isolate pure retrieved-context signal alongside combined performance, our empirical benchmark evaluates standalone text ($Z_i$), baseline covariates ($X_i$), and combined features ($(X_i,Z_i)$); because $\tau_{\mathcal{X}}$, $\tau_{\mathcal{X},\mathcal{Z}}$, and $\Gamma_{\pi,\phi}$ are unobserved, we score out-of-sample prioritization using the rank-weighted functional below. 

\paragraph{The RATE ratio as a verification metric.} Because CATEs have no per-unit ground truth, we evaluate whether a representation separates high- from low-benefit units out of sample using a rank-based functional. Given out-of-fold priority scores from a CATE estimator, the rank-weighted average treatment effect ($\mathrm{RATE}$) weights units by how strongly they are prioritized, with AUTOC weighting as primary and Qini weighting as a robustness check \citep{yadlowsky2025evaluating}. With standard error $\widehat{\mathrm{se}}_{\mathrm{cl}}$ clustered at the randomization-unit level, the RATE ratio is
\begin{equation}
\text{RATE ratio} = \widehat{\mathrm{RATE}} \,\big/\,
\widehat{\mathrm{se}}_{\mathrm{cl}}\!\left(\widehat{\mathrm{RATE}}\right).
\label{eq:rate}
\end{equation}
Under the regularity conditions of \citet{yadlowsky2025evaluating}, the RATE ratio~\eqref{eq:rate} is asymptotically standard normal under $H_0:\mathrm{RATE}=0$; we write AUTOC/SE and Qini/SE for its AUTOC- and Qini-weighted instantiations.

\section{The benchmark protocol: three verification layers}
\label{sec:protocol}

\paragraph{Structural decomposition: a two-channel bound.} Even when retrieved documents predate $t_i$, post-treatment information can enter $Z_i$ through post-$t_i$ pretraining records in the encoder weights $\phi$ or through the agentic retrieval policy. Theorem~\ref{thm:main} bounds the pretraining term and keeps the summary term explicit. Let $\Psi$ denote a post-$t_i$ quantity correlated with the potential outcomes---the realized outcome or, in an encoder-leakage audit, a post-cutoff event label (such as a subject's later death year or a subsequent earthquake). The encoder weights are produced by a pretraining algorithm
\[
  \phi \;=\; \Alg(\Cpre, \Gpost, R),
\]
where $\Cpre$ is the pre-$t_i$ portion of the pretraining corpus, $\Gpost = (G_1,\ldots,G_k)$ consists of $k$ post-$t_i$ training records referencing the event (leaving the $t_i$-dependence of $k$ and $\Psi$ implicit), and $R$ is the algorithm's internal randomness. This is a closed structural model: every data source or selection rule through which $\Psi$ can affect training must appear among $(\Cpre,\Gpost,R)$. Three assumptions---representation construction, a closed training model, and per-record KL stability with constant $\beta$---are stated in Appendix~\ref{app:proofs}.

\begin{theorem}[Two-channel information bound]\label{thm:main}
Under Assumptions~\ref{a:iso}--\ref{a:stab},
\[
  \I(Z_i; \Psi \mid \Cpre)
  \;\le\; k\beta+\I(S_i;\Psi\mid\phi,\Cpre).
\]
Moreover, $\I(\phi;\Psi\mid\Cpre)\le\I(\phi;\Gpost\mid\Cpre)\le k\beta$. For $k=0$, the product $k\beta$ is interpreted as zero.
\end{theorem}

Theorem~\ref{thm:main} motivates the three-layer structure below (proofs and a differential-privacy corollary in Appendix~\ref{app:proofs}). In the retrieval channel $\I(S_i;\Psi\mid\phi,\Cpre)$, excluding post-$t_i$ documents prevents direct outcome contamination---motivating Layer~1's temporal filter (Section~\ref{sec:layer1}), though timestamps alone cannot bound policy weights $\Theta_{\pi_i}$ (Remark~\ref{rem:retrieval}). Because admissible pre-$t_i$ summaries still lack unit-level counterfactuals and risk search overfitting---while leaked summaries would inflate out-of-sample scores---Layer~2 pairs Layer~1 with cross-fitted RATE validation (Section~\ref{sec:layer2}). In the weight channel, post-treatment influence is bounded by $k\beta$, the number of event-referencing pretraining records times per-record KL sensitivity. Because $\beta$ is unknown for off-the-shelf encoders, Theorem~\ref{thm:main} bounds this channel only structurally; in Layer~3, we probe it empirically with vintaged encoders (Section~\ref{sec:layer3}) and calibrate it at exact $\beta$ values under Gaussian noise (Section~\ref{sec:case}).

\subsection{Layer 1: retrieval admissibility}
\label{sec:layer1}

\paragraph{Temporal provenance filter.} For each located, dated experimental unit, both workflows withhold treatment assignments and outcomes from the agent while enforcing pre-$t_i$ date bounds at retrieval (Appendix~\ref{app:agent}). \emph{Session-isolated retrieval} applies ex-post \emph{no-post} filtering to per-context web and archive searches, dropping explicitly post-cutoff items while retaining undated excerpts. The \emph{workspace teammate} (GrokBot) compiles one shared corpus per study and enforces stricter \emph{known-pre} filtering, retaining only sources with verified pre-intervention dates and no outcome mentions. Even under strict date filtering, however, document timestamps cannot rule out post-$t_i$ knowledge in the retrieval agent's own weights $\Theta_{\pi_i}$.

\subsection{Layer 2: heterogeneity validation}
\label{sec:layer2}

\paragraph{Cross-fitted RATE scoring.} Once candidate representations pass Layer~1's temporal filters, Layer~2 evaluates whether they resolve genuine out-of-sample treatment-effect heterogeneity. To isolate the contribution of each representation---comparing tabular covariates $X_i$, agentic text $Z_i$, and concatenations such as $(X_i,Z_i)$---we train causal forests \citep{wager2018estimation,athey2019generalized} on identical samples, cross-fitting folds, and hyperparameters. From the resulting out-of-fold priority scores and shared doubly robust evaluation scores, we compute the AUTOC-weighted RATE ratio~\eqref{eq:rate} with standard errors clustered at the randomization-unit level, and report Qini-weighted ratios in robustness checks.

\paragraph{RATIO: a proposed Tier-4 search specification.} Whereas our benchmark evaluates fixed Tier-3 workflows, Tier-4 pipeline search (Equation~\eqref{eq:agentic-levels}) must guard against adaptive overfitting; we therefore discuss---though leave as a secondary extension for future investigation---a sample-split search loop, RATIO (Appendix~\ref{app:ratio}). Subject to the admissibility constraints, an agent proposes covariate subsets, feature maps, retrieval policies, encoders, and estimator settings across fixed cluster-preserving development splits, using validation-subset RATE ratios solely as an optimization signal. Within a fixed search budget, RATIO selects the candidate with the highest mean validation ratio, refits it once on the full development sample, and evaluates the frozen pipeline once on the held-out test split.

Yet, even strict sample splitting only helps prevent overfitting; it cannot detect post-treatment knowledge memorized in pretrained encoder weights, which requires Layer~3.

\subsection{Layer 3: encoder admissibility}
\label{sec:layer3}

\paragraph{Chronological vintage probes.} Passing Layer~1 guarantees only that the input text predates $t_i$; the encoder $g_{\phi}$ can still inject post-$t_i$ facts memorized during pretraining. We probe this channel with chronologically vintaged encoders trained on chronologically bounded corpora \citep{he2025chronologically}: hold the input text fixed and always pre-event, vary only the encoder vintage, and ask whether later vintages recover future facts from the same historical text better than earlier vintages. Because successive vintages change the entire intervening corpus and optimization trajectory alongside $\Gpost$, vintage comparisons serve as observational diagnostics. The \emph{death-year probe} predicts subjects' later death years from biography snapshots written while the subjects were alive. The \emph{earthquake probe} embeds pre-disaster Wikipedia pages for struck and matched control cities using encoders trained before and after each quake. Weight leakage would pull struck cities' embeddings---more than controls'---toward disaster terms in post-quake encoders.

\paragraph{Calibration instruments.} Observational probes alone cannot distinguish a leakage-free encoder from an insensitive probe, nor attribute a detected shift to specific post-$t_i$ records; two calibration instruments help close these gaps. First, to verify that our embedding and linear probe can recover a future label when it enters the text---a check sensitive to token pooling and input length---each death-year probe pairs clean historical biographies with positive controls that prepend or append the subject's death year explicitly. Second, to isolate single-record parameter memorization from broader corpus drift between vintages---and to make the per-record stability constant $\beta$ in Theorem~\ref{thm:main} finite and analytically exact---we pair observational probes with a Gaussian-perturbed adapter experiment that also tests whether post-hoc weight noise can erase memorized facts without destroying fluency.

\section{Empirical verification: workflows, encoders, and leakage audits}
\label{sec:case}

\subsection{Anchor experiment: Uganda YOP}

\begin{figure}[t]
\centering
\includegraphics[width=0.95\textwidth]{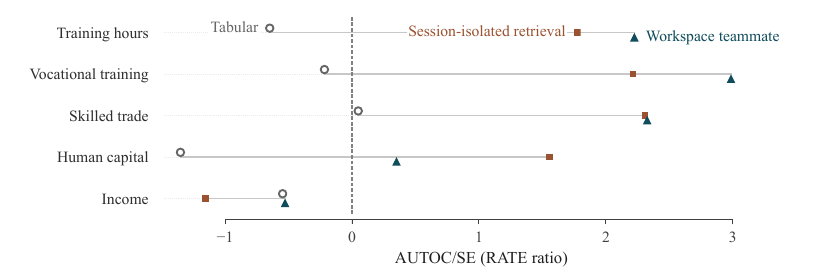}
\caption{\textbf{Uganda YOP matched RATE ratios.} Pooled out-of-fold AUTOC/SE for tabular covariates, session-isolated retrieval, and workspace-teammate (GrokBot) text. All representations share identical samples and saved folds (Qini/SE in Table~\ref{tab:uganda}).}
\label{fig:uganda}
\end{figure}

We first apply Layers~1 and~2 to our anchor field experiment, Uganda's 2008 Youth Opportunities Program (YOP), which randomized unconditional cash grants (approximately US\$400 per member) to young-adult groups across \UgandaClusterN{} design clusters (\UgandaNMin{}--\UgandaNMax{} matched individuals \citep{blattman2014generating,blattman2020long}). Across \UgandaOutcomeCount{} intention-to-treat outcomes, we compare baseline survey covariates against text from session-isolated retrieval, which ran an independent session per parish, and from the workspace teammate (GrokBot), which re-verified candidate URLs from the session-isolated run alongside pre-2008 district development and humanitarian reports. Using the small OpenAI encoder, we extract \HeadlinePCN{} principal components across the study's unlabeled text embeddings---a low-dimensional compression chosen to curb split dilution on place-level embeddings in axis-aligned trees (Appendix~\ref{app:agent})---and fit honest causal forests (\HeadlineTreeN{} trees) across shared \OuterFoldN{}-fold cross-fitting splits. Temporally filtered local text outperforms baseline survey covariates on all four skill and human-capital outcomes---training hours, vocational training, skilled trade, and human capital (Fig.~\ref{fig:uganda}). On income, however, every representation yields a negative AUTOC/SE ratio ($\UgandaOriginalIncomeRatio{}$ for session-isolated text, $\UgandaGrokIncomeRatio{}$ for GrokBot, and $\UgandaTabularIncome{}$ for tabular covariates). To test whether a persistent workspace teammate systematically improves upon session-isolated retrieval beyond this single trial, we next compare both pipelines across the full multi-study battery.

\subsection{Multi-study workflow and encoder sensitivity}

Across the full battery of \BatteryStudyCount{} datasets and \BatteryContrastCount{} eligible treatment--outcome contrasts, the GrokBot workspace teammate achieves a higher within-study mean AUTOC/SE in \GrokTextHigherStudies{} datasets for text alone and \GrokCombinedHigherStudies{} for combined tabular-plus-text features (Table~\ref{tab:workflow}). Across context-level and area-pooled variants of text and combined features (\WorkflowComparisonN{} paired AUTOC tests; context-level means in Table~\ref{tab:workflow}), \WorkflowSignificantN{} differences pass Holm multiplicity adjustment: \WorkflowPositiveN{} favor the workspace teammate and \WorkflowNegativeN{} favor session-isolated retrieval (Appendix~\ref{app:workflow}). GrokBot gains most when local context lives in structured administrative PDFs: in the Kenya digital-ID trial, parsing county census and development tables lifts mean text AUTOC/SE from \BatteryTableAP{} to \BatteryTableAQ{} (and combined features from \BatteryTableAR{} to \BatteryTableAS{}, passing Holm adjustment on tax registration). It also wins Holm-significant contrasts in direct air capture (area-pooled text) and the footbridge arm of trolley dilemmas. Conversely, session-isolated retrieval tends to win when strict \emph{known-pre} temporal filtering leaves GrokBot with sparse governance reports---in civic honesty on area-pooled text (context-level text AUTOC/SE likewise drops from $\BatteryTableCR{}$ to $\BatteryTableCS{}$). Because the two workflows vary base model, tooling, search scope, and date filtering together, this comparison evaluates complete acquisition pipelines.

\begin{table}[t]
\centering
\small
\setlength{\tabcolsep}{3pt}
\caption{\textbf{Session-isolated retrieval and workspace-teammate (GrokBot) workflow comparison.} Within-study means of contrast-specific AUTOC/SE ratios. The shared tabular benchmark uses identical samples and evaluation scores.}
\label{tab:workflow}
\input{FiguresMake/generated/battery_table}
\end{table}

Our primary causal-forest benchmark fixes a single encoder compressed to \HeadlinePCN{} principal components fitted across unlabeled study embeddings---a tree-specific parsimony choice supported by pilot causal-forest width sweeps where wider PCA budgets diluted place-level signal (Appendix~\ref{app:agent}). To test whether \HeadlinePCN{} components remain optimal once direct $L_2$ regularization replaces axis-aligned tree splitting---and to test sensitivity to encoder choice---we next run an R-learner analysis \citep{nie2021quasi} that confines all scaling, PCA, nuisance estimation, and penalty tuning strictly to each outer training fold (Appendix~\ref{app:encoder}). Evaluating uncompressed (pre-PCA) embeddings (\SmallEmbeddingDim{}--\MPNetEmbeddingDim{} dimensions) alongside an in-fold \HeadlinePCN{}-PC small-encoder control, we compare the large OpenAI encoder and MPNet against the small OpenAI baseline. Across \EncoderTestN{} paired AUTOC comparisons on uncompressed embeddings, \EncoderSignificantN{} remain significant after Holm--Bonferroni adjustment, split evenly between favoring the alternative encoder (\EncoderPositiveN{}) and favoring the small OpenAI baseline (\EncoderNegativeN{}); comparing uncompressed against in-fold \HeadlinePCN{}-component small-encoder embeddings likewise yields \CompressionSignificantN{} significant differences across \CompressionTestN{} tests (\CompressionPositiveN{} favoring uncompressed embeddings, \CompressionNegativeN{} favoring PCA). Compression to \HeadlinePCN{} PCs is thus an estimator-matched choice for causal forests rather than universally dominant under $L_2$ regularization, and both encoder choice and compression materially shift CATE rankings in either direction. Taken together, the \BatteryStudyCount{}-study RCT analyses show that date-filtered agentic context can capture treatment-effect heterogeneity missed by baseline surveys, yet downstream CATE rankings remain sensitive to both retrieval workflow and pretrained encoder choice---motivating Layer~3's direct audits of encoder memorization below.

\subsection{Encoder leakage audits: vintage probes and \texorpdfstring{$\beta$}{beta}-calibration}

\begin{figure}[t]
\centering
\includegraphics[width=0.87\textwidth]{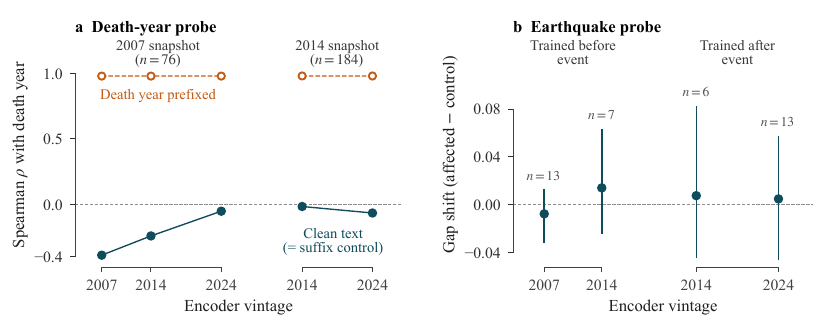}
\caption{\textbf{Vintage comparisons and explicit-label controls.} \textbf{a}, Spearman correlations ($\rho$) with actual death year for clean biographies (identical to the suffix control) and biographies with the death year prepended, across cohorts ($n=\DeathEarlyN{}$ and $n=\DeathLateN{}$) and encoder vintages. \textbf{b}, Earthquake gap shifts with 95\% pair-bootstrap intervals for encoders trained before or after each event; $n$ counts city pairs. \QuakePairN{} pairs are nested in \QuakeEventN{} events.}
\label{fig:leakage}
\end{figure}

\paragraph{Observational vintage probes and pooling controls.} To test whether pretrained encoders inject memorized post-event facts into temporally clean text, we apply Layer~3 using ChronoGPT encoders with 2007, 2014, and 2024 training cutoffs \citep{he2025chronologically} while holding historical inputs fixed (Fig.~\ref{fig:leakage}; Appendix~\ref{app:audit}). In the death-year probe on 2007 ($n=\DeathEarlyN{}$) and 2014 ($n=\DeathLateN{}$) biography snapshots of then-living subjects, clean text yields negative rank correlations with death year across all vintages (Spearman $\rho \in [\DeathCleanRhoMin{}, \DeathCleanRhoMax{}]$); binary late-death AUC rises across vintages in the smaller 2007 cohort (\DeathDetailATG{} to \DeathDetailAUZ{}) but falls from \DeathTableASV{} (2014 vintage) to below chance at \DeathTableASU{} (2024 vintage) in the larger 2014 cohort (Table~\ref{tab:death-detail}). Explicit-label controls clarify both the rationale and scope of our primary pooling window: to equalize sequence length across articles and prevent short prepended control labels from washing out over long texts, our primary pooling rule averages the first \DeathPoolTokenN{} token states (Appendix~\ref{app:audit}). Because ChronoGPT uses left-to-right causal attention, hidden states at tokens $1$--$\DeathPoolTokenN{}$ cannot attend to any token after position~$\DeathPoolTokenN{}$. Prepending the death year at token~$1$ therefore enters the $\DeathPoolTokenN{}$-token average directly and yields near-perfect recovery ($\rho \ge \DeathPrefixRhoMin{}$), whereas appending it at the end of the article (past token~$\DeathPoolTokenN{}$) cannot alter the first $\DeathPoolTokenN{}$ hidden states at all, leaving the pooled embedding identical to that of clean text---illustrating how pooling rules govern what a leakage probe can detect. To test place-level shocks that mirror geographic RCTs, we embed pre-disaster pages for \QuakePairN{} affected/control city pairs (\QuakeEventN{} earthquakes). For the 2024 encoder, the post-earthquake shift is \QuakeShift{} (95\% CI $[\QuakeLower{},\ \QuakeUpper{}]$); as detailed in Appendix~\ref{app:audit}, confidence intervals still include zero after subtracting flood or city-term control shifts and when pooling the final \QuakePoolTokenN{} tokens or encoding city names alone.

\begin{figure}[t]
\centering
\resizebox{0.88\textwidth}{!}{\input{FiguresMake/fig_beta_setup}}
\caption{\textbf{Testing single-record weight memorization under Gaussian weight noise.} Two corpora share pre-2026 background texts ($\Cpre$) and differ in one post-cutoff record ($k=1$) stating that a fictional subject died in 2031 (World $\Psi=0$, weights $\mu_0$) or 2041 (World $\Psi=1$, weights $\mu_1$). Adding Gaussian noise $\mathcal{N}(\mathbf{0},\sigma_{\phi}^2\mathbf{I})$ sets an exact one-record KL bound $\beta=\|\mu_1-\mu_0\|_2^2/(2\sigma_{\phi}^2)$; encoding a clean 2026 biography $S_i$ (containing neither year) then tests whether $Z_i=g_{\phi}(S_i)$ still predicts the memorized year and how much held-out fluency degrades as noise rises.}
\label{fig:beta-setup}
\end{figure}
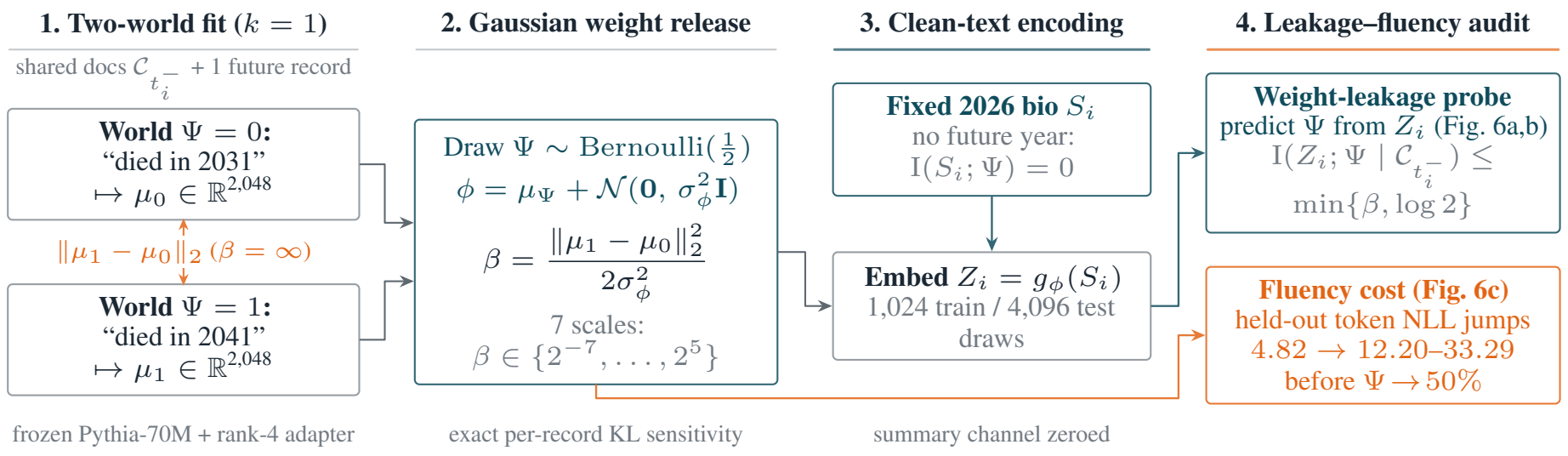

\begin{figure}[t]
\centering
\includegraphics[width=0.87\textwidth]{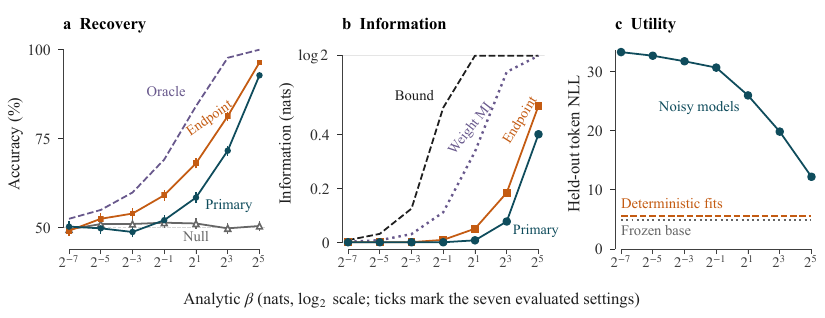}
\caption{\textbf{Controlled Gaussian calibration.} \textbf{a}, Accuracy of the primary and endpoint-projection probes (pointwise \BetaConfidenceLevel{}\% Wilson intervals), plotted alongside the Bayes-optimal weight classifier (analytic weight oracle) and the primary probe under the identical-mean null control. Because the identical-mean null control has true $\beta=0$ (which cannot appear on the log axis), each null run is plotted at the $\beta$ coordinate of the signal run that shares its Gaussian noise variance $\sigma_{\phi}^2$. \textbf{b}, The entropy-capped upper bound, exact Gaussian weight-channel mutual information, and simultaneous probe information lower bounds ($\I_{\mathrm{lo}}$). \textbf{c}, Mean held-out token negative log-likelihood across \BetaUtilityN{} noisy model draws per setting; horizontal lines give the frozen-base value and the mean of the two deterministic fits.}
\label{fig:beta-calibration}
\end{figure}

\paragraph{Controlled Gaussian $\beta$-calibration.} Because observational vintage probes cannot isolate single-record memorization from broader drift in the training corpus or yield a known per-record KL stability bound $\beta$ for the pretraining process, we complement them with a restricted training procedure in which $\beta$ is analytically exact (Fig.~\ref{fig:beta-setup}). Starting from a frozen open-weight Pythia-70M encoder, we deterministically fit a rank-\BetaAdapterRank{} residual adapter with \BetaParameterN{} trainable coefficients to each of two corpora that differ only in one record assigning a fictitious person a death year of 2031 or 2041, yielding distinct noiseless adapter vectors $\mu_0$ and $\mu_1$. Deterministic fitting alone produces point-mass parameter laws with infinite KL divergence ($\beta=\infty$), leaving Theorem~\ref{thm:main} vacuous. We therefore inject isotropic Gaussian output noise $\mathcal{N}(\mathbf{0}, \sigma_{\phi}^2 \mathbf{I})$ around $\mu_0$ and $\mu_1$ for two purposes: first, it converts training into a randomized Gaussian mechanism with closed-form stability $\beta=\|\mu_1-\mu_0\|_2^2/(2\sigma_{\phi}^2)$ so Theorem~\ref{thm:main} can be calibrated across exact $\beta$ regimes; second, it tests whether post-hoc weight perturbation---a post-training weight-perturbation mitigation (Corollary~\ref{cor:dp})---can mask memorized post-treatment facts without retraining. Because the evaluation biography (mean-pooled across all input tokens) is identical across all randomized models and contains neither death year, the summary term vanishes; with $k=1$ and a balanced draw between the two candidate years ($\Psi \sim \mathrm{Bernoulli}(1/2)$, capping binary entropy at $\log 2$ nats), Theorem~\ref{thm:main} gives $\I(Z_i;\Psi\mid\Cpre)\le\min\{\beta,\log 2\}$. At each of \BetaNoiseLevelN{} noise scales, we fit a standardized linear probe to embeddings from \BetaTrainN{} randomized models and test it on \BetaTestN{} independently drawn models.

Empirical recovery tracks the theoretical bound closely, but suppressing weight leakage via post-hoc noise severely degrades model fluency (Fig.~\ref{fig:beta-calibration}; Appendix~\ref{app:beta-calibration}). Primary-probe accuracy drops from \BetaAccuracyHigh{}\% at $\beta=2^5$ ($\BetaHigh{}$ nats) and \BetaAccuracyNext{}\% at $\beta=2^3$ ($\BetaNext{}$) to \BetaAccuracyMid{}\% at $\beta=2^{-1}$ ($\BetaMid{}$), reaching chance once $\beta \le 2^{-3}$; at $\beta=2^{-7}$, primary accuracy is \BetaAccuracyLow{}\% (matching the \BetaNullAccuracyMin{}--\BetaNullAccuracyMax{}\% null controls) and endpoint projection also falls to chance (Table~\ref{tab:beta-calibration}). Across these same noise scales, however, held-out token negative log-likelihood jumps from \BetaBaseUtility{} (frozen base; $5.52$ for the noiseless fitted adapters) to \BetaUtilityMin{} already at $\beta=2^5\ (\BetaHigh{})$---where probe recovery is still \BetaAccuracyHigh{}\%---and \BetaUtilityMax{} at $\beta=2^{-7}$, showing that isotropic weight noise destroys fluency well before memorized future facts disappear. Moreover, at $\beta=2^5\ (\BetaHigh{})$ the probe recovers the future year with \BetaUnrelatedHigh{}\% accuracy even from an unrelated biography, showing that parameter leakage shifts embeddings globally beyond text naming the training subject.

\section{Discussion and limitations}
\label{sec:discussion}

\paragraph{Takeaway.} Across \BatteryStudyCount{} studies, neither retrieval architecture dominates CATE ranking, and temporal filters at retrieval cannot confirm admissibility alone: pretrained encoders shift CATE rankings and can retain memorized future facts, while post-hoc weight noise degrades language-model fluency well before erasing that memorization.

\paragraph{Beyond causal inference.} The protocol's three verification layers apply beyond causal inference whenever agents construct features under an information constraint. Temporal provenance filtering applies to any forecasting, nowcasting, or prediction task where labels must not leak into inputs. Paired, cluster-robust RATE testing likewise applies whenever researchers rank units with agent-generated signals. Vintage probes transfer wherever a pretrained model's corpus may overlap with evaluation targets, echoing contamination checks for language-model benchmarks.

\paragraph{Limitations.} A hosted workspace teammate relies on unpinned model weights and a shared corpus whose sources still require direct audit. Our observational leakage probes cover specific encoders, token-pooling rules, and historical events, while the controlled $\beta$ experiment studies one invented fact under noisy adapter perturbation.

\paragraph{Data and code availability.} The encoder-probe suite is available at \url{https://huggingface.co/datasets/cjerzak/AgenticHeterogeneity}. Tabular baselines can be recomputed from publicly archived trial data \citep{blattman2020long}; related code and data for the Uganda RCT are at \url{https://huggingface.co/datasets/cjerzak/ImageHeterogeneity}.

\bibliographystyle{unsrtnat}
\bibliography{mybib}

\clearpage
\appendix

\section{Two-channel information bound: assumptions, proofs, and scope}
\label{app:proofs}

All random elements are assumed to take values in standard Borel spaces, so regular conditional laws exist. Conditional statements are understood for $P_{\Cpre}$-almost every corpus $c$, and mutual information and KL divergence use natural logarithms. Let $\Psi$ denote a post-$t_i$ quantity correlated with the potential outcomes, such as the realized outcome or a post-cutoff event label (e.g., a subject's death year or a subsequent earthquake). Temporal document filtering may exclude future documents, but $\Psi$ can remain predictable from past text or enter through either the summary policy's parameters or the encoder's weights.

\begin{assumption}[Representation construction]\label{a:iso}
$Z_i=g_{\phi}(S_i)$ is a measurable deterministic function of $(\phi,S_i)$. No independence between $S_i$, $\phi$, and $\Psi$ is imposed.
\end{assumption}

\begin{assumption}[Closed training model]\label{a:markov}
$\phi=\Alg(\Cpre,\Gpost,R)$ and $R\indep\Psi\mid(\Cpre,\Gpost)$. In particular, no omitted training input or selection rule may depend on $\Psi$ once $(\Cpre,\Gpost)$ is fixed.
\end{assumption}

\begin{assumption}[Per-record KL stability]\label{a:stab}
The records $G_1,\ldots,G_k$ in $\Gpost=(G_1,\ldots,G_k)$ are mutually independent conditional on $\Cpre$, and there exists $\beta<\infty$ such that, for $P_{\Cpre}$-almost every $c$, every admissible realization $u=(u_1,\ldots,u_k)$ of $\Gpost$, every index $j\le k$, and every admissible replacement record $u'$,
\[
  \KL{P(\phi \mid \Cpre=c, \Gpost=u)}{P(\phi \mid \Cpre=c, \Gpost=u^{(j \to u')})} \;\le\; \beta,
\]
where $u^{(j \to u')}$ denotes $u$ with its $j$-th coordinate replaced by $u'$.
\end{assumption}

The constant $\beta$ is a uniform upper bound on the relative-entropy influence of one training record. Conditional independence of $G_1,\ldots,G_k$ given $\Cpre$ is a substantive assumption on how post-$t_i$ records are generated. Likewise, a finite $\beta$ is not automatic: deterministic training can produce disjoint point masses and infinite KL.

\begin{lemma}[Conditional data processing]\label{l:dpi}
Under Assumption~\ref{a:iso},
\[
\I(Z_i;\Psi\mid\Cpre)
\le \I(\phi,S_i;\Psi\mid\Cpre)
=\I(\phi;\Psi\mid\Cpre)+\I(S_i;\Psi\mid\phi,\Cpre).
\]
\end{lemma}

The final term is the open summary/retrieval channel. It may contain ordinary predictive pretreatment signal as well as information introduced by a post-period retrieval or summarization policy.

\begin{lemma}[Fact-to-group information bound]\label{l:coarsen}
Under Assumption~\ref{a:markov}, $\;\I(\phi; \Psi \mid \Cpre) \le \I(\phi; \Gpost \mid \Cpre)$.
\end{lemma}

\begin{lemma}[Group information under KL stability]\label{l:mi}
Under Assumption~\ref{a:stab}, $\;\I(\phi; \Gpost \mid \Cpre) \le k\beta$.
\end{lemma}

\begin{proof}[Proof of Lemma~\ref{l:dpi}]
Conditional on $\Cpre$, $Z_i=g_{\phi}(S_i)$ is a deterministic measurable function of $(\phi,S_i)$. Conditional data processing and the chain rule give
\[
  \I(Z_i;\Psi\mid\Cpre)
  \le \I(\phi,S_i;\Psi\mid\Cpre)
  =\I(\phi;\Psi\mid\Cpre)+\I(S_i;\Psi\mid\phi,\Cpre).
\]
No independence is used.
\end{proof}

\begin{proof}[Proof of Lemma~\ref{l:coarsen}]
If $\I(\phi;\Gpost\mid\Cpre)=\infty$, the claim is immediate. Otherwise, for $P_{\Cpre}$-almost every $c$ the first term below is finite, and we may expand $\I(\phi;\Psi,\Gpost\mid c)$ by the chain rule in two ways:
\[
  \I(\phi; \Gpost \mid c) + \I(\phi; \Psi \mid \Gpost, c)
  \;=\; \I(\phi; \Psi \mid c) + \I(\phi; \Gpost \mid \Psi, c).
\]
Assumption~\ref{a:markov} gives $\phi\indep\Psi\mid(\Cpre,\Gpost)$, so the second term on the left vanishes. The second term on the right is nonnegative. Rearranging, $\I(\phi; \Psi \mid c) \le \I(\phi; \Gpost \mid c)$. Average over $c$.
\end{proof}

\begin{proof}[Proof of Lemma~\ref{l:mi}]
If $k=0$, both sides are zero. Otherwise, fix a value $\Cpre=c$ outside the exceptional null set and suppress it. By the chain rule, $\I(\phi; \Gpost) = \sum_{j=1}^{k} \I(\phi; G_j \mid G_{<j})$ with $G_{<j} = (G_1,\ldots,G_{j-1})$. Fix $j$ and condition on $G_{<j}=u_{<j}$. Write $Q_{u_j} \coloneqq P(\phi \mid G_{<j}=u_{<j}, G_j=u_j)$ and take $G'_j$ to be an independent copy of $G_j$; mutual independence of the coordinates then gives $P(\phi \mid G_{<j}=u_{<j}) = \mathbb{E}[Q_{G'_j}]$, so by the definition of conditional mutual information, convexity of relative entropy in its second argument, and Jensen's inequality,
\[
  \I(\phi; G_j \mid G_{<j}=u_{<j})
  \;=\; \mathbb{E}\!\left[\KL{Q_{G_j}}{\mathbb{E}\!\left[Q_{G'_j}\right]}\right]
  \;\le\; \mathbb{E}\!\left[\KL{Q_{G_j}}{Q_{G'_j}}\right].
\]
For any realizations $u_j,u'$ of $G_j$, both $Q_{u_j}$ and $Q_{u'}$ are mixtures of $P(\phi \mid \Gpost=(u_{<j},\cdot,G_{>j}))$ over $G_{>j}$ with the same mixing law, again by independence. Joint convexity of relative entropy therefore gives
\[
  \KL{Q_{u_j}}{Q_{u'}}
  \;\le\; \mathbb{E}\!\left[\KL{P\bigl(\phi \mid \Gpost=(u_{<j},u_j,G_{>j})\bigr)}{P\bigl(\phi \mid \Gpost=(u_{<j},u',G_{>j})\bigr)}\right]
  \;\le\; \beta
\]
by Assumption~\ref{a:stab}. Averaging over $G_{<j}$ shows each of the $k$ summands is at most $\beta$; averaging over $c$ completes the proof.
\end{proof}

\begin{proof}[Proof of Theorem~\ref{thm:main}]
Apply Lemma~\ref{l:dpi} and bound its first summand using Lemmas~\ref{l:coarsen} and~\ref{l:mi}. The displayed subsidiary chain is exactly Lemmas~\ref{l:coarsen} and~\ref{l:mi}.
\end{proof}

The encoder-only bound $\I(Z_i;\Psi\mid\Cpre)\le k\beta$ follows only under the additional condition $\I(S_i;\Psi\mid\phi,\Cpre)=0$, equivalently $S_i\indep\Psi\mid(\phi,\Cpre)$ when regular conditional laws are used. Temporal document filtering alone does not imply this condition.

\begin{corollary}[Differential privacy as a special case]\label{cor:dp}
Under Assumptions~\ref{a:iso} and~\ref{a:markov}, suppose the conditional-independence clause of Assumption~\ref{a:stab} holds and $\Alg$ is pure $\varepsilon_0$-differentially private under bounded (replacement) adjacency of one training record while $\Cpre$ is held fixed. Then the KL-stability clause holds with $\beta=\varepsilon_0(e^{\varepsilon_0}-1)$, and
\[
  \I(Z_i;\Psi\mid\Cpre)
  \le k\,\varepsilon_0\bigl(e^{\varepsilon_0}-1\bigr)
  +\I(S_i;\Psi\mid\phi,\Cpre).
\]
\end{corollary}

\begin{proof}
For the output laws $P,Q$ on replacement-neighboring datasets, pure differential privacy in both directions gives $e^{-\varepsilon_0}\le dP/dQ\le e^{\varepsilon_0}$. Hence
\[
\KL{P}{Q}\le \KL{P}{Q}+\KL{Q}{P}
=\int (dP-dQ)\log\!\left(\frac{dP}{dQ}\right)
\le\varepsilon_0(e^{\varepsilon_0}-1).
\]
The last inequality uses $|\log(dP/dQ)|\le\varepsilon_0$ and $\int|dP-dQ|\le e^{\varepsilon_0}-1$. Hence Assumption~\ref{a:stab} holds with $\beta=\varepsilon_0(e^{\varepsilon_0}-1)$, and the bound follows from Theorem~\ref{thm:main}. Approximate privacy is not covered by this likelihood-ratio argument, and add/remove adjacency would require the corresponding replacement or group-privacy constants.
\end{proof}

\begin{remark}[Conditioning on $\Cpre$]\label{rem:conditioning}
The quantity $\I(Z_i;\Psi\mid\Cpre)$ measures residual association after conditioning on the chosen corpus. When the relevant information is finite,
\[
\I(Z_i;\Psi\mid\Cpre)
=\I(Z_i,\Cpre;\Psi)-\I(\Cpre;\Psi).
\]
Because conditioning can increase mutual information, $\I(Z_i;\Psi\mid\Cpre)$ can exceed $\I(Z_i;\Psi)$; interpreting it as post-treatment leakage therefore requires a sufficiency or causal model for $\Cpre$.
\end{remark}

\begin{remark}[Residual retrieval channel]\label{rem:retrieval}
Theorem~\ref{thm:main} retains the retrieval and summarization channel as $\I(S_i;\Psi\mid\phi,\Cpre)$. The term can mix legitimate predictive pretreatment content with leakage through $\Theta_{\pi_i}$, and timestamps alone do not separate them. Such a separation would require a baseline representation $\mathcal{B}_i$ capturing the admissible pretreatment information in $\Filt$ together with an explicit causal or conditional-independence model relative to $\mathcal{B}_i$. If the policy is itself trained, an analogous stability analysis may bound the contribution of its weights; it cannot simply be assumed to vanish.
\end{remark}

\section{Study battery, retrieval architectures, and primary estimation}
\label{app:agent}

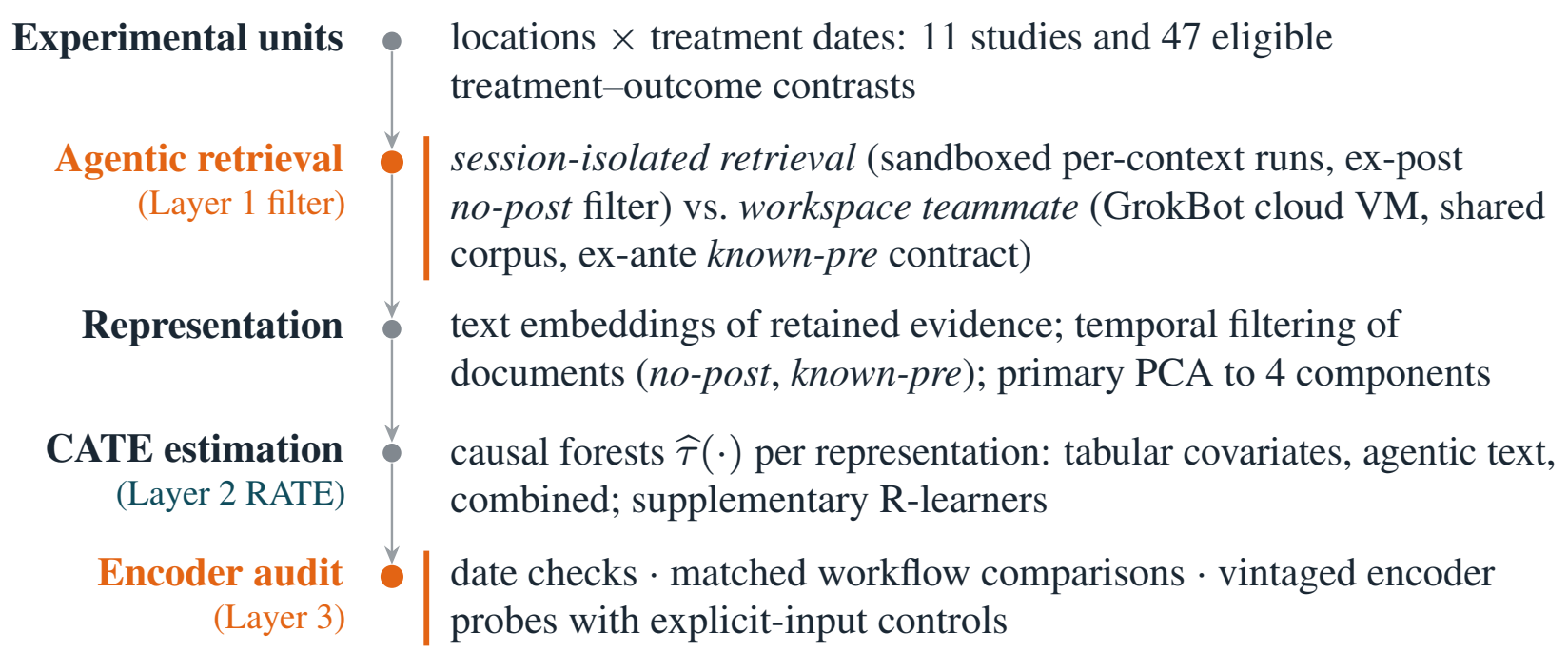
\begin{figure}[t]
\centering
\resizebox{0.82\textwidth}{!}{%
\begin{tikzpicture}[
  x=1cm, y=1cm,
  >={Stealth[length=3.6pt,width=3.0pt]},
  stitle/.style={font=\footnotesize\bfseries, text=ink, align=flush right, text width=2.85cm, inner sep=0pt},
  gtitle/.style={stitle, text=accent},
  sdesc/.style={font=\footnotesize, text=ink, align=flush left, text width=7.65cm, inner sep=0pt},
  sdot/.style={circle, fill=ink!55, inner sep=1.3pt},
  gdot/.style={circle, fill=accent, inner sep=1.6pt},
  sarr/.style={->, line width=0.45pt, draw=ink!45}
]
\useasboundingbox (0,-4.32) rectangle (11.455,0.28);

% Stage 1: Experimental units
\node[stitle, anchor=north east] at (2.95, 0.12) {Experimental units};
\node[sdot] (d1) at (3.28, -0.02) {};
\node[sdesc, anchor=north west]  at (3.68, 0.12)
  {locations $\times$ treatment dates: \BatteryStudyCount{} studies and
   \BatteryContrastCount{} eligible treatment--outcome contrasts};

% Stage 2: Agentic retrieval (Orange verification stage -- Layer 1)
\node[gtitle, anchor=north east] at (2.95, -0.72) {Agentic retrieval\\[-1pt]{\scriptsize\normalfont\color{accent}(Layer 1 filter)}};
\node[gdot] (d2) at (3.28, -0.86) {};
\draw[accent, line width=1.1pt] (3.52, -0.68) -- (3.52, -1.68);
\node[sdesc, anchor=north west]  at (3.68, -0.72)
  {\emph{session-isolated retrieval} (sandboxed per-context runs, ex-post \emph{no-post} filter)
   vs.\ \emph{workspace teammate} (GrokBot cloud VM, shared corpus, ex-ante \emph{known-pre} contract)};

% Stage 3: Representation
\node[stitle, anchor=north east] at (2.95, -1.88) {Representation};
\node[sdot] (d3) at (3.28, -2.02) {};
\node[sdesc, anchor=north west]  at (3.68, -1.88)
  {text embeddings of retained evidence; temporal filtering of documents
   (\emph{no-post}, \emph{known-pre}); primary PCA to \HeadlinePCN{} components};

% Stage 4: CATE estimation (Layer 2)
\node[stitle, anchor=north east] at (2.95, -2.74) {CATE estimation\\[-1pt]{\scriptsize\normalfont\color{deep}(Layer 2 RATE)}};
\node[sdot] (d4) at (3.28, -2.88) {};
\node[sdesc, anchor=north west]  at (3.68, -2.74)
  {causal forests $\widehat{\tau}(\cdot)$ per representation: tabular covariates,
   agentic text, combined; supplementary R-learners};

% Stage 5: Encoder audit (Orange verification stage -- Layer 3)
\node[gtitle, anchor=north east] at (2.95, -3.60) {Encoder audit\\[-1pt]{\scriptsize\normalfont\color{accent}(Layer 3)}};
\node[gdot] (d5) at (3.28, -3.74) {};
\draw[accent, line width=1.1pt] (3.52, -3.56) -- (3.52, -4.22);
\node[sdesc, anchor=north west]  at (3.68, -3.60)
  {date checks $\cdot$ matched workflow comparisons $\cdot$
   vintaged encoder probes with explicit-input controls};

% Connecting flow spine
\draw[sarr] (d1) -- (d2);
\draw[sarr] (d2) -- (d3);
\draw[sarr] (d3) -- (d4);
\draw[sarr] (d4) -- (d5);
\end{tikzpicture}%
}
\caption{\textbf{The agentic pipeline.} Located, dated experimental units pass through agentic retrieval and temporal filtering (Layer~1)---instantiated either via session-isolated per-context runs or via a workspace teammate (GrokBot)---followed by text embedding, cross-fitted CATE estimation and RATE scoring (Layer~2), and encoder-weight leakage audits (Layer~3). Orange stages mark retrieval-provenance and encoder-leakage checks.}
\label{fig:pipeline}
\end{figure}

\paragraph{Study battery and matched samples.} Uganda estimation samples contain \UgandaNMin{}--\UgandaNMax{} observations in \UgandaClusterN{} design clusters, depending on the outcome. Each contrast uses identical units, weights, folds, propensities, and evaluation scores across representations. Beyond Uganda YOP, our \BatteryStudyCount{}-dataset battery (Table~\ref{tab:composition}) comprises four located preference-conjoint experiments---Kenya digital ID \citep{garbe2025who}, US direct air capture siting \citep{scottbuechler2024communities}, US just energy transition policy \citep{gazmararian2024fossil}, and US COVID-19 booster uptake \citep{raman2022covid}, with each randomized attribute defining a separate contrast---alongside six public field, clinical, and behavioral experiments: Bangladesh masks \citep{abaluck2022masking}, holiday advertisements \citep{breza2021advertising}, IST-1 stroke \citep{sandercock2011database}, Moral Machine \citep{awad2018moral}, trolley dilemmas \citep{awad2020universals}, and civic honesty \citep{cohn2019honesty}.

\begin{table}[ht]
\centering
\small
\caption{\textbf{Completed battery and eligible samples.} Ranges span the eligible treatment--outcome contrasts within each dataset.}
\label{tab:composition}
\input{FiguresMake/generated/composition_table}
\end{table}

\paragraph{Retrieval workflows and provenance.} Under \emph{session-isolated retrieval} (Table~\ref{tab:workflow-arch}), an automated worker pool spawns an ephemeral, read-only CLI session per study context (with repeated retrieval draws on the six multi-site public trials), calling a pinned \texttt{gpt-5.4-mini} deployment with bounded web-search, Wayback, and lightweight page-reading tools and checking the returned factor, verbatim-evidence, URL, and recorded-date fields ex post. Under the \emph{workspace teammate} workflow, xAI's GrokBot on \texttt{grok.com} runs in a persistent cloud virtual machine with a browser, PDF-parsing, and Python tooling. For each study, our benchmark harness provides GrokBot with a self-contained handoff package containing canonical place registries, contrast definitions, and candidate URLs from the session-isolated run while withholding treatment assignments and evaluation outcomes until the corpus is frozen. Within its VM, GrokBot builds a shared study corpus---filtering out-of-window candidates, parsing dated administrative PDFs, querying archived web snapshots and bibliographic records, and mapping verbatim excerpts across places---before an offline validator verifies \emph{known-pre} publication intervals and excludes study-outcome keywords.

Both workflows have provenance limits. In GrokBot's run logs, we verify that outcomes were withheld during retrieval, though xAI does not expose an exact checkpoint identifier. Because GrokBot's handoff included candidate URLs from the session-isolated run, the two corpora are not independent retrieval samples; replication also differs (repeated per-context draws in six studies under session-isolated retrieval versus one study-wide pass under GrokBot), and in IST-1---where patient recruitment began in 1991 and ran through 1996 prior to the trial's May 1997 publication---we tightened the date cutoff from May 1997 (session-isolated) to December 31, 1990 (GrokBot) so all sources predate patient enrollment.

\begin{table}[ht]
\centering
\footnotesize
\setlength{\tabcolsep}{4pt}
\renewcommand{\arraystretch}{1.12}
\caption{\textbf{Architectural and governance comparison of the two Tier-3 agent workflows.}}
\label{tab:workflow-arch}
\begin{tabular}{@{}>{\raggedright\arraybackslash}p{2.25cm}>{\raggedright\arraybackslash}p{5.35cm}>{\raggedright\arraybackslash}p{5.65cm}@{}}
\toprule
Dimension & Session-isolated retrieval & Workspace teammate (GrokBot) \\
\midrule
Compute runtime & Ephemeral, read-only CLI sessions calling a pinned \texttt{gpt-5.4-mini} deployment & Persistent cloud Linux VM on \texttt{grok.com} with browser, PDF-parsing, and Python tools \\
Retrieval topology ($\pi_i$) & Isolated per-context sessions (16-turn, 24-call budget; repeated draws on six public trials); no cross-context memory & Stateful study-wide pass over the canonical place registry; shared source cache and scripted place mapping \\
Source \& archive tooling & Web search, Wayback lookup, and lightweight HTML/short-PDF text reader (no headless JS rendering or bulk PDF table parsing) & Browser rendering, bulk PDF table parsing, Wayback snapshots, PubMed/GitHub APIs, and candidate-URL re-verification \\
Temporal \& leakage filter & Prompt window plus ex-post \emph{no-post} date filter (retains undated rows) and published-trial self-citation regex filter & Ex-ante offline validator: \emph{known-pre} date-interval bounds and study-outcome keyword exclusion \\
Outcome blinding & Worker inputs omit assignments and outcomes by script construction & Handoff package withholds evaluation outcomes until pre-evaluation corpus freeze \\
Governance trade-off & Pinned model deployment and fixed per-context prompt, but cross-session search noise and undated excerpts & Zero undated items and richer document parsing, but unpinned model weights and shared candidate URLs \\
\bottomrule
\end{tabular}
\end{table}

\paragraph{Pilot prompt-optimization check.} During pilot development, we also tested an outcome-guided prompt loop in which an optimizer model iteratively rewrote retrieval prompts to maximize training-fold RATE ratios while holding out evaluation folds. Selecting the best of $M$ candidate prompts raised in-sample scores but failed to generalize out of sample---motivating strict outcome blinding during Tier-3 retrieval and a dedicated untouched test split in RATIO (Appendix~\ref{app:ratio}).

\paragraph{Primary representations and causal-forest estimation.} Across the end-to-end verification pipeline (Fig.~\ref{fig:pipeline}), retained evidence paragraphs are concatenated per context and embedded using \texttt{text-\allowbreak embedding-\allowbreak 3-\allowbreak small}. The original \emph{no-post} filter retains undated sources; GrokBot's \emph{known-pre} requires a parseable date on or before the configured cutoff. Study-specific PCA uses all unlabeled context vectors and retains \HeadlinePCN{} components to suit axis-aligned tree splitting. Because text embeddings vary only across study contexts (as few as \CompositionDN{}--\CompositionEG{} places; Table~\ref{tab:composition}) and reproducible signal concentrates in the first component, wider PCA budgets ($16$, $64$, or $128$ components) diluted splits on noisy trailing axes and never improved context-level AUTOC/SE in pilot sweeps. Area-pooled variants average context embeddings within administrative areas before PCA. Honest causal forests \citep{wager2018estimation,athey2019generalized} use \HeadlineTreeN{} trees and the saved \OuterFoldN{}-fold place-component partitions. Design clusters follow the assignment or repeated-observation unit appropriate to each dataset.

\begin{table}[htbp]
\centering
\small
\setlength{\tabcolsep}{3pt}
\caption{\textbf{Uganda YOP RATE ratios by outcome and representation.} AUTOC/SE and Qini/SE ratios use \RateBootstrapN{} shared resamples. Session-isolated retrieval and workspace-teammate (GrokBot) workflows retain their respective filters, with a common embedding and forest specification across outcomes.}
\label{tab:uganda}
\resizebox{\textwidth}{!}{\input{FiguresMake/generated/uganda_table}}
\end{table}

We pool one out-of-fold priority score per observation to estimate the RATE, using AUTOC weighting for the primary analysis and Qini weighting as a sensitivity check \citep{yadlowsky2025evaluating}. Priority scores are held fixed, all representations are evaluated against the same doubly robust scores, and standard errors and paired differences are computed from \RateBootstrapN{} cluster-level half-sample bootstrap draws (following the \texttt{grf} RATE procedure) shared across the compared representations. All \WorkflowComparisonN{} primary workflow comparisons (text, combined, and their area-pooled variants) form one Holm family. Paired RATE differences between text and tabular representations are adjusted for multiplicity as a separate Holm family. As a sensitivity check, we re-estimate standard errors and paired tests with clustering by place component rather than by the study's design cluster. For study-level summaries, we report unweighted means of contrast-specific ratios \citep{alam2025decision}.

\section{Layer 3 leakage probes and Gaussian calibration}
\label{app:audit}

\subsection{Observational vintage probes}
\label{app:vintage-probes}

\paragraph{Chronologically vintaged encoders.} The probes use ChronoGPT \citep{he2025chronologically}, a family of language models trained with chronologically bounded corpora. The vintages used are the chronologically consistent instruct checkpoints with cutoffs 2007-12-31, 2014-12-31, and 2024-12-31 (\texttt{manelalab/\allowbreak chrono-\allowbreak gpt-\allowbreak instruct-\allowbreak v1-\allowbreak \{20071231, 20141231, 20241231\}}). All probes hold the input text fixed and historical, varying the encoder vintage.

\paragraph{Death-year probe.} Historical Wikipedia biographies concern people alive at snapshot time who later died (2007 cohort: \DeathEarlyN{}; 2014 cohort: \DeathLateN{}). Our primary probe configuration averages the first \DeathPoolTokenN{} hidden states with a \DeathMaxTokenN{}-token input limit. This opening-\DeathPoolTokenN{}-token window was selected during pilot calibration because full biographies almost always exceed \DeathMaxTokenN{} tokens---causing end-of-article suffix controls to be truncated before encoding even under last-\DeathPoolTokenN{}-token pooling---while averaging the final \DeathPoolTokenN{} states of the \DeathMaxTokenN{}-token window diluted a prepended death-year label across more than 100 intervening tokens, leaving the linear probe no more accurate than on clean text (zero MAE or rank-correlation gain). Averaging the first \DeathPoolTokenN{} causal states equalizes sequence length across snapshots and aligns the pooled representation with the prefix control, at the cost of ignoring text past token~\DeathPoolTokenN{} (and, on unstripped Wikipedia revisions, occasionally pooling leading infobox or hatnote markup alongside or ahead of the subject's name); suffix-control scores therefore match clean text identically in this configuration (Tables~\ref{tab:probes} and~\ref{tab:death-detail}).

\begin{table}[ht]
\centering
\small
\caption{\textbf{Death-year vintage probe summary.} Bracketed intervals span the available vintages within each cohort. MAE is in years; window accuracy is balanced accuracy; AUC concerns late-death classification.}
\label{tab:probes}
\resizebox{\textwidth}{!}{\input{FiguresMake/generated/death_table}}
\end{table}

\begin{table}[ht]
\centering
\footnotesize
\setlength{\tabcolsep}{4pt}
\caption{\textbf{Death-year vintage probe results by snapshot cohort and encoder vintage.} Suffix-control scores equal the corresponding Clean rows identically under first-\DeathPoolTokenN{}-token causal pooling and are omitted.}
\label{tab:death-detail}
\input{FiguresMake/generated/death_detail}
\end{table}

\paragraph{Earthquake probe.} Starting from \QuakeSourceEventN{} candidate earthquakes (\QuakeSourcePairN{} candidate city pairs, \QuakePageN{} historical city pages), restricting to complete pre-event affected/control pairs yields \QuakePairN{} comparisons across \QuakeEventN{} events. For encoder $g_{\phi}$ and concept $\kappa$, the pair gap is $\operatorname{gap}_{\phi}(\kappa)=\operatorname{dist}_{\phi}(\mathrm{affected},\kappa)-\operatorname{dist}_{\phi}(\mathrm{control},\kappa)$, where $\operatorname{dist}_{\phi}$ denotes cosine distance between unit-normalized embeddings. We define a pair's post-event shift by subtracting its mean pre-event gap from its post-event gap; because smaller cosine distance indicates closer embedding proximity, a negative post-event shift means that post-quake encoders pull struck places closer to earthquake terms relative to matched controls. To test specificity to earthquakes, we subtract the corresponding shift for a control concept (floods or generic city terms) from the earthquake-concept shift. Matching the death-year audit, our primary probe configuration averages the first \QuakePoolTokenN{} hidden states with a \QuakeMaxTokenN{}-token input limit. Because unstripped Wikipedia city pages often open with infobox markup inside the first \QuakePoolTokenN{} tokens, we also test last-\QuakePoolTokenN{}-token pooling across the \QuakeMaxTokenN{}-token window and city-name-only inputs ($32$-token cap). Both city-level checks yield 2024 post-event shifts whose 95\% confidence intervals include zero, whereas bare region-name prompts (e.g., province or prefecture names by which major quakes are commonly named) pull struck regions closer to earthquake terms (negative gap shift).

\subsection{Controlled Gaussian \texorpdfstring{$\beta$}{beta}-calibration}
\label{app:beta-calibration}
In the controlled Pythia-70M calibration, deterministic full-batch Adam yields two fixed rank-\BetaAdapterRank{} final-block MLP up-projection vectors ($\mu_0$ and $\mu_1$, with \BetaParameterN{} trainable coefficients and all other weights frozen) whose point-mass parameter laws would have infinite KL divergence without randomization; adding common isotropic Gaussian output perturbations $\mathcal{N}(\mathbf{0}, \sigma_{\phi}^2 \mathbf{I})$ induces an exact two-world stability constant $\beta=\|\mu_1-\mu_0\|_2^2/(2\sigma_{\phi}^2)$ while testing whether adding Gaussian weight noise after training can mask memorized post-treatment facts without retraining the encoder. Unlike the ChronoGPT prefix pooling rule, each Pythia embedding averages final hidden states across the entire fixed input biography. As an upper-reference linear benchmark, the noiseless-direction projection projects noisy embeddings onto the exact noiseless vector difference $g_{\mu_1}(S_i)-g_{\mu_0}(S_i)$ between both known training worlds, isolating whether signal missed by the primary linear probe reflects weight noise rather than high-dimensional probe estimation variance. Its classification threshold is fitted only on training draws. To form the information lower bounds, we compute a one-sided exact binomial lower confidence bound on probe accuracy---Bonferroni-adjusted across all \BetaFamilyN{} combinations of probe type, biography input, signal versus null-control condition, and noise scale---and substitute that bound into the binary Fano inequality (Table~\ref{tab:beta-calibration}; Fig.~\ref{fig:beta-calibration}).

\begin{table}[htbp]
\centering
\footnotesize
\setlength{\tabcolsep}{3pt}
\renewcommand{\arraystretch}{1.08}
\caption{\textbf{Target-biography probe calibration.} Accuracy is in percent with pointwise \BetaConfidenceLevel{}\% Wilson intervals; AUC is the area under the ROC curve. $\I_{\mathrm{lo}}$ is the simultaneous \BetaConfidenceLevel{}\% information lower bound in nats. The endpoint projection uses both noiseless training-world embeddings. All \BetaNoiseLevelN{} noise settings use \BetaTrainN{} training and \BetaTestN{} independent test model releases.}
\label{tab:beta-calibration}
\input{FiguresMake/generated/beta_table}
\end{table}

\section{Contrast-level workflow results and encoder sensitivity}
\label{app:workflow}

\subsection{Contrast-level RATE ratios and paired workflow tests}
\label{app:contrast-results}

Across all \BatteryContrastCount{} contrasts, we compute primary RATE ratios per representation (Table~\ref{tab:contrast-detail}) and paired workspace-teammate (GrokBot) minus session-isolated retrieval AUTOC differences adjusted within a single \WorkflowComparisonN{}-test Holm family. Table~\ref{tab:paired-detail} reports all 14 comparisons in this family with Holm-adjusted $p < 1$; the remaining 174 comparisons yield $\text{Holm } p = 1$ (in studies where administrative areas match retrieval contexts, such as ZIP-level holiday advertisements, context-level and area-pooled specifications coincide).

\begin{table}[htbp]
\centering
\footnotesize
\setlength{\tabcolsep}{4pt}
\renewcommand{\arraystretch}{1.08}
\caption{\textbf{Paired workspace-teammate minus session-isolated retrieval AUTOC differences ($\text{Holm } p < 1$).} Pointwise 95\% normal intervals use \RateBootstrapN{} shared cluster-bootstrap draws across the \WorkflowComparisonN{}-test workflow multiplicity family; all remaining 174 comparisons yield $\text{Holm } p = 1$.}
\label{tab:paired-detail}
\resizebox{\textwidth}{!}{\input{FiguresMake/generated/paired_table}}
\end{table}

\subsection{Uncompressed embeddings and penalized R-learner estimation}
\label{app:encoder}

As an exploratory post-hoc sensitivity check added after inspecting the primary causal-forest results, we evaluate uncompressed (pre-PCA) embeddings with $L_2$-penalized R-learners. We re-embed the complete retained evidence with the small and large OpenAI encoders (\texttt{text-\allowbreak embedding-\allowbreak 3-\allowbreak small} and \texttt{text-\allowbreak embedding-\allowbreak 3-\allowbreak large}, both evaluated at \SmallEmbeddingDim{} Matryoshka dimensions) and \texttt{sentence-\allowbreak transformers/\allowbreak all-\allowbreak mpnet-\allowbreak base-\allowbreak v2} at \MPNetEmbeddingDim{} dimensions \citep{allmpnetv2}. OpenAI inputs use nonoverlapping byte-bounded chunks; MPNet uses complete token windows. Splitting documents into chunks avoids truncation, so all input text contributes to the representation; we then combine chunk embeddings via length-weighted averaging. Because architecture, dimension, and chunking vary together across models, we compare complete encoding pipelines.

The CATE estimator is an R-learner \citep{nie2021quasi} parameterizing $\tau(\tilde{Z}_i) = \alpha + \tilde{Z}_i^\top \beta_{\tau}$ on standardized features $\tilde{Z}_i$ and minimizing weighted R-loss $\bigl(\sum_i w_i\bigr)^{-1}\sum_i w_i\bigl[(Y_i-\widehat{m}_i)-(W_i-\widehat{e}_i)(\alpha+\tilde{Z}_i^\top\beta_{\tau})\bigr]^2 + \lambda\|\beta_{\tau}\|_2^2$. Here $\lambda \ge 0$ is an $L_2$ \emph{slope penalty} applied only to the heterogeneity slope vector $\beta_{\tau}$ while leaving the CATE intercept $\alpha$ unpenalized: because uncompressed embeddings span \SmallEmbeddingDim{}--\MPNetEmbeddingDim{} dimensions across as few as \CompositionDN{} unique places, penalizing $\beta_{\tau}$ prevents high-dimensional overfitting and shrinks weak heterogeneity toward a constant treatment effect ($\beta_{\tau}\to\mathbf{0}$) without shrinking the average effect $\alpha$. Within each saved outer fold, nested grouped validation chooses the slope penalty $\lambda$ over a logarithmic grid from $10^{-4}$ to $10^6$, in half-log steps. Nuisance fits, imputation, scaling, and the small-encoder \HeadlinePCN{}-component control use only their designated training partitions. Outcome nuisances use fixed-penalty $L_2$-regularized linear regression; propensities use known design probabilities or training-only logistic regression. Missing text is represented by zeros and an availability indicator. All specifications share the frozen evaluation samples and scores; CATE predictions are not clipped. Unlike unsupervised PCA compression, fitting an R-learner on uncompressed embeddings regularizes the representation directly under causal R-loss.

Of \EncoderTestN{} paired encoder contrasts, \EncoderSignificantN{} remain significant after Holm--Bonferroni adjustment: \EncoderPositiveN{} favor the alternative encoder and \EncoderNegativeN{} favor the small OpenAI encoder. Across \CompressionTestN{} paired tests comparing uncompressed (\SmallEmbeddingDim{}-dimensional) embeddings from the small OpenAI encoder against its in-fold \HeadlinePCN{}-component PCA control, \CompressionSignificantN{} remain significant after adjustment: \CompressionPositiveN{} favor the uncompressed embeddings and \CompressionNegativeN{} favor the \HeadlinePCN{}-component control. In the separate Holm family testing paired differences over the tabular baseline, \SupplementIncrementSignificantN{} of \SupplementIncrementTestN{} comparisons are significant (\SupplementIncrementPositiveN{} positive, \SupplementIncrementNegativeN{} negative). Slope penalties reach a grid boundary in \RidgeBoundaryN{} of \RidgeFitN{} outer fits, primarily reflecting shrinkage toward a constant CATE ($\lambda=10^6$) when fold-level heterogeneity is weak (study-level means in Table~\ref{tab:encoder-detail}).

\begin{landscape}
\scriptsize
\setlength{\LTpre}{5pt}
\setlength{\LTpost}{5pt}
\renewcommand{\arraystretch}{0.96}
\input{FiguresMake/generated/contrast_table}
\input{FiguresMake/generated/encoder_table}
\end{landscape}

\section{RATIO: adaptive pipeline optimization}
\label{app:ratio}

RATIO implements the Tier-4 capability in Equation~\eqref{eq:agentic-levels} by using RATE-ratio probes to revise the entire analysis pipeline. Let
\[
\eta=(J,h_2,\pi,\phi,h_3,\lambda)
\]
collect covariate selection, feature construction, the retrieval-and-summarization specification (including $\Theta_{\pi}$), encoder weights, feature integration, and CATE-estimator settings, respectively. Write $Z_i(\eta)$ for the representation induced by this configuration. The treatment, outcome, RATE weighting, and admissibility requirements are fixed before search; the agent proposes configurations within these constraints.

\paragraph{Development splits and validation scores.} Split the sample by randomization cluster into development indices $\mathcal{I}_{\mathrm{dev}}$ and an untouched test set $\mathcal{I}_{\mathrm{test}}$. Partition the development sample into $B$ validation subsets $\mathcal{V}_1,\ldots,\mathcal{V}_B$, retaining the same splits across candidates. For each configuration $\eta$ and subset $\mathcal{V}_b$, fit the priority rule, preprocessing, and nuisance models on $\mathcal{I}_{\mathrm{dev}}\setminus\mathcal{V}_b$, and let $\widehat{\mathrm{RATE}}_{\mathcal{V}_b}(\eta)$ denote the resulting RATE estimate on $\mathcal{V}_b$. The subset probe is
\begin{equation}
r_b(\eta)=
\frac{\widehat{\mathrm{RATE}}_{\mathcal{V}_b}(\eta)}
{\widehat{\mathrm{se}}_{\mathrm{cl}}\!\left(
\widehat{\mathrm{RATE}}_{\mathcal{V}_b}(\eta)\right)},
\qquad b=1,\ldots,B,
\label{eq:ratio-probe}
\end{equation}
where the standard error clusters at the randomization-unit level.

\paragraph{Agent feedback and selection.} Let $\omega$ be the agent's proposal policy and let its history after $m$ evaluated candidates be
\begin{equation}
\mathcal{H}_m=
\left\{\big(\eta_s,r_1(\eta_s),\ldots,r_B(\eta_s)\big)\right\}_{s=1}^{m},
\qquad
\eta_{m+1}=\omega(\mathcal{H}_m).
\label{eq:ratio-feedback}
\end{equation}
The feedback can trigger changes to any component of $\eta$, including new retrieval or feature construction. After evaluating $M$ candidates within a prespecified search budget, select
\begin{equation}
\widehat{\eta}\in
\operatorname*{arg\,max}_{\eta\in\{\eta_1,\ldots,\eta_M\}}
\frac{1}{B}\sum_{b=1}^{B}r_b(\eta),
\label{eq:ratio-selection}
\end{equation}
using a prespecified tie rule. Because search inspects validation subsets repeatedly, this mean serves only as an optimization objective, not a standard-normal test statistic.

\paragraph{Final evaluation and scope.} Freeze the selected specification and refit it on $\mathcal{I}_{\mathrm{dev}}$. Evaluate its resulting priority scores once on $\mathcal{I}_{\mathrm{test}}$, reporting one RATE estimate and a cluster-robust standard error under the benchmark's regularity conditions. The agent receives no test-set feedback during search, and the reported evaluation does not trigger another optimization round. A comparison with a prespecified baseline uses a paired, covariance-aware RATE contrast, as in Section~\ref{sec:protocol}.

For an admissible selected pipeline, the information-gain identity in Equation~\eqref{eq:agentic-gain} applies conditional on its development data; maximizing subset RATE ratios does not imply maximizing $\Gamma_{\pi,\phi}$. An optional extension probes subpopulations defined by pretreatment covariates. Any selected subgroup is frozen before test-set evaluation, and the resulting RATE pertains to that subgroup.

\end{document}

%% file: FiguresMake/generated/numbers.tex
\newcommand{\HeadlinePCN}{4}%
\newcommand{\OuterFoldN}{5}%
\newcommand{\SmallEmbeddingDim}{256}%
\newcommand{\LargeEmbeddingDim}{256}%
\newcommand{\MPNetEmbeddingDim}{768}%
\newcommand{\BatteryStudyCount}{11}%
\newcommand{\BatteryContrastCount}{47}%
\newcommand{\GrokTextHigherStudies}{6}%
\newcommand{\GrokCombinedHigherStudies}{7}%
\newcommand{\RateBootstrapN}{2000}%
\newcommand{\HeadlineTreeN}{1000}%
\newcommand{\WorkflowComparisonN}{188}%
\newcommand{\WorkflowSignificantN}{7}%
\newcommand{\WorkflowPositiveN}{5}%
\newcommand{\WorkflowNegativeN}{2}%
\newcommand{\UgandaNMin}{1710}%
\newcommand{\UgandaNMax}{1715}%
\newcommand{\UgandaClusterN}{439}%
\newcommand{\UgandaOutcomeCount}{5}%
\newcommand{\UgandaOriginalIncomeRatio}{-1.15}%
\newcommand{\UgandaGrokIncomeRatio}{-0.53}%
\newcommand{\UgandaTabularIncome}{-0.55}%
\newcommand{\BatteryTableAH}{5}%
\newcommand{\BatteryTableAI}{-0.54}%
\newcommand{\BatteryTableAJ}{1.34}%
\newcommand{\BatteryTableAK}{1.47}%
\newcommand{\BatteryTableAL}{0.97}%
\newcommand{\BatteryTableAM}{1.16}%
\newcommand{\BatteryTableAN}{5}%
\newcommand{\BatteryTableAO}{3.18}%
\newcommand{\BatteryTableAP}{0.78}%
\newcommand{\BatteryTableAQ}{2.50}%
\newcommand{\BatteryTableAR}{0.99}%
\newcommand{\BatteryTableAS}{3.43}%
\newcommand{\BatteryTableAT}{7}%
\newcommand{\BatteryTableAU}{-0.52}%
\newcommand{\BatteryTableAV}{0.87}%
\newcommand{\BatteryTableAW}{0.50}%
\newcommand{\BatteryTableAX}{0.55}%
\newcommand{\BatteryTableAY}{0.19}%
\newcommand{\BatteryTableAZ}{7}%
\newcommand{\BatteryTableBA}{-0.82}%
\newcommand{\BatteryTableBB}{-0.51}%
\newcommand{\BatteryTableBC}{-0.15}%
\newcommand{\BatteryTableBD}{-0.78}%
\newcommand{\BatteryTableBE}{-0.71}%
\newcommand{\BatteryTableBF}{5}%
\newcommand{\BatteryTableBG}{-1.08}%
\newcommand{\BatteryTableBH}{-0.26}%
\newcommand{\BatteryTableBI}{0.44}%
\newcommand{\BatteryTableBJ}{-0.35}%
\newcommand{\BatteryTableBK}{0.04}%
\newcommand{\BatteryTableBL}{2}%
\newcommand{\BatteryTableBM}{0.48}%
\newcommand{\BatteryTableBN}{-0.63}%
\newcommand{\BatteryTableBO}{-0.16}%
\newcommand{\BatteryTableBP}{0.53}%
\newcommand{\BatteryTableBQ}{0.33}%
\newcommand{\BatteryTableBR}{2}%
\newcommand{\BatteryTableBS}{-0.08}%
\newcommand{\BatteryTableBT}{0.65}%
\newcommand{\BatteryTableBU}{0.06}%
\newcommand{\BatteryTableBV}{1.18}%
\newcommand{\BatteryTableBW}{1.39}%
\newcommand{\BatteryTableBX}{4}%
\newcommand{\BatteryTableBY}{-0.17}%
\newcommand{\BatteryTableBZ}{-0.26}%
\newcommand{\BatteryTableCA}{-0.27}%
\newcommand{\BatteryTableCB}{-0.63}%
\newcommand{\BatteryTableCC}{-0.38}%
\newcommand{\BatteryTableCD}{7}%
\newcommand{\BatteryTableCE}{1.60}%
\newcommand{\BatteryTableCF}{1.94}%
\newcommand{\BatteryTableCG}{1.37}%
\newcommand{\BatteryTableCH}{4.61}%
\newcommand{\BatteryTableCI}{4.85}%
\newcommand{\BatteryTableCJ}{2}%
\newcommand{\BatteryTableCK}{8.87}%
\newcommand{\BatteryTableCL}{3.95}%
\newcommand{\BatteryTableCM}{5.13}%
\newcommand{\BatteryTableCN}{10.32}%
\newcommand{\BatteryTableCO}{9.88}%
\newcommand{\BatteryTableCP}{1}%
\newcommand{\BatteryTableCQ}{3.28}%
\newcommand{\BatteryTableCR}{0.25}%
\newcommand{\BatteryTableCS}{-2.44}%
\newcommand{\BatteryTableCT}{3.91}%
\newcommand{\BatteryTableCU}{2.49}%
\newcommand{\CompositionCV}{5}%
\newcommand{\CompositionCW}{1,710}%
\newcommand{\CompositionCX}{1,715}%
\newcommand{\CompositionCY}{296}%
\newcommand{\CompositionCZ}{5}%
\newcommand{\CompositionDA}{14,076}%
\newcommand{\CompositionDB}{1,224}%
\newcommand{\CompositionDC}{7}%
\newcommand{\CompositionDD}{29,068}%
\newcommand{\CompositionDE}{494}%
\newcommand{\CompositionDF}{7}%
\newcommand{\CompositionDG}{9,220}%
\newcommand{\CompositionDH}{407}%
\newcommand{\CompositionDI}{5}%
\newcommand{\CompositionDJ}{9,655}%
\newcommand{\CompositionDK}{579}%
\newcommand{\CompositionDL}{2}%
\newcommand{\CompositionDM}{595}%
\newcommand{\CompositionDN}{34}%
\newcommand{\CompositionDO}{2}%
\newcommand{\CompositionDP}{6,717}%
\newcommand{\CompositionDQ}{6,773}%
\newcommand{\CompositionDR}{754}%
\newcommand{\CompositionDS}{788}%
\newcommand{\CompositionDT}{4}%
\newcommand{\CompositionDU}{19,285}%
\newcommand{\CompositionDV}{19,435}%
\newcommand{\CompositionDW}{36}%
\newcommand{\CompositionDX}{7}%
\newcommand{\CompositionDY}{10,618}%
\newcommand{\CompositionDZ}{116,950}%
\newcommand{\CompositionEA}{71}%
\newcommand{\CompositionEB}{2}%
\newcommand{\CompositionEC}{135,996}%
\newcommand{\CompositionED}{41}%
\newcommand{\CompositionEE}{1}%
\newcommand{\CompositionEF}{16,104}%
\newcommand{\CompositionEG}{40}%
\newcommand{\UgandaTableEH}{-0.65}%
\newcommand{\UgandaTableEI}{1.77}%
\newcommand{\UgandaTableEJ}{2.22}%
\newcommand{\UgandaTableEK}{0.64}%
\newcommand{\UgandaTableEL}{2.81}%
\newcommand{\UgandaTableEM}{2.67}%
\newcommand{\UgandaTableEN}{-0.22}%
\newcommand{\UgandaTableEO}{2.21}%
\newcommand{\UgandaTableEP}{2.98}%
\newcommand{\UgandaTableEQ}{1.22}%
\newcommand{\UgandaTableER}{3.23}%
\newcommand{\UgandaTableES}{3.83}%
\newcommand{\UgandaTableET}{0.05}%
\newcommand{\UgandaTableEU}{2.31}%
\newcommand{\UgandaTableEV}{2.32}%
\newcommand{\UgandaTableEW}{-0.73}%
\newcommand{\UgandaTableEX}{2.21}%
\newcommand{\UgandaTableEY}{2.10}%
\newcommand{\UgandaTableEZ}{-1.35}%
\newcommand{\UgandaTableFA}{1.55}%
\newcommand{\UgandaTableFB}{0.35}%
\newcommand{\UgandaTableFC}{-2.07}%
\newcommand{\UgandaTableFD}{1.44}%
\newcommand{\UgandaTableFE}{1.07}%
\newcommand{\UgandaTableFF}{-0.55}%
\newcommand{\UgandaTableFG}{-1.15}%
\newcommand{\UgandaTableFH}{-0.53}%
\newcommand{\UgandaTableFI}{0.25}%
\newcommand{\UgandaTableFJ}{-1.14}%
\newcommand{\UgandaTableFK}{-0.08}%
\newcommand{\ContrastTableFL}{1.81}%
\newcommand{\ContrastTableFM}{-0.54}%
\newcommand{\ContrastTableFN}{-0.88}%
\newcommand{\ContrastTableFO}{1.35}%
\newcommand{\ContrastTableFP}{1.05}%
\newcommand{\ContrastTableFQ}{-0.86}%
\newcommand{\ContrastTableFR}{-0.73}%
\newcommand{\ContrastTableFS}{0.57}%
\newcommand{\ContrastTableFT}{-0.29}%
\newcommand{\ContrastTableFU}{-0.40}%
\newcommand{\ContrastTableFV}{-0.01}%
\newcommand{\ContrastTableFW}{0.28}%
\newcommand{\ContrastTableFX}{0.34}%
\newcommand{\ContrastTableFY}{0.57}%
\newcommand{\ContrastTableFZ}{-0.05}%
\newcommand{\ContrastTableGA}{-2.85}%
\newcommand{\ContrastTableGB}{-0.49}%
\newcommand{\ContrastTableGC}{0.24}%
\newcommand{\ContrastTableGD}{-2.44}%
\newcommand{\ContrastTableGE}{-1.50}%
\newcommand{\ContrastTableGF}{-1.53}%
\newcommand{\ContrastTableGG}{-0.67}%
\newcommand{\ContrastTableGH}{0.66}%
\newcommand{\ContrastTableGI}{0.70}%
\newcommand{\ContrastTableGJ}{0.29}%
\newcommand{\ContrastTableGK}{-1.85}%
\newcommand{\ContrastTableGL}{-0.57}%
\newcommand{\ContrastTableGM}{-0.61}%
\newcommand{\ContrastTableGN}{-0.75}%
\newcommand{\ContrastTableGO}{-0.43}%
\newcommand{\ContrastTableGP}{0.82}%
\newcommand{\ContrastTableGQ}{0.16}%
\newcommand{\ContrastTableGR}{1.57}%
\newcommand{\ContrastTableGS}{0.15}%
\newcommand{\ContrastTableGT}{1.88}%
\newcommand{\ContrastTableGU}{3.28}%
\newcommand{\ContrastTableGV}{0.25}%
\newcommand{\ContrastTableGW}{-2.44}%
\newcommand{\ContrastTableGX}{3.91}%
\newcommand{\ContrastTableGY}{2.49}%
\newcommand{\ContrastTableGZ}{-1.49}%
\newcommand{\ContrastTableHA}{0.29}%
\newcommand{\ContrastTableHB}{0.60}%
\newcommand{\ContrastTableHC}{0.48}%
\newcommand{\ContrastTableHD}{-0.22}%
\newcommand{\ContrastTableHE}{-0.51}%
\newcommand{\ContrastTableHF}{2.60}%
\newcommand{\ContrastTableHG}{2.78}%
\newcommand{\ContrastTableHH}{0.87}%
\newcommand{\ContrastTableHI}{2.30}%
\newcommand{\ContrastTableHJ}{0.24}%
\newcommand{\ContrastTableHK}{0.49}%
\newcommand{\ContrastTableHL}{-0.99}%
\newcommand{\ContrastTableHM}{0.18}%
\newcommand{\ContrastTableHN}{-0.57}%
\newcommand{\ContrastTableHO}{0.42}%
\newcommand{\ContrastTableHP}{2.44}%
\newcommand{\ContrastTableHQ}{2.36}%
\newcommand{\ContrastTableHR}{2.47}%
\newcommand{\ContrastTableHS}{2.15}%
\newcommand{\ContrastTableHT}{-2.15}%
\newcommand{\ContrastTableHU}{-0.06}%
\newcommand{\ContrastTableHV}{-2.04}%
\newcommand{\ContrastTableHW}{-0.67}%
\newcommand{\ContrastTableHX}{-2.92}%
\newcommand{\ContrastTableHY}{0.63}%
\newcommand{\ContrastTableHZ}{-0.17}%
\newcommand{\ContrastTableIA}{1.68}%
\newcommand{\ContrastTableIB}{0.17}%
\newcommand{\ContrastTableIC}{1.80}%
\newcommand{\ContrastTableID}{-0.76}%
\newcommand{\ContrastTableIE}{0.49}%
\newcommand{\ContrastTableIF}{-0.87}%
\newcommand{\ContrastTableIG}{0.31}%
\newcommand{\ContrastTableIH}{-1.23}%
\newcommand{\ContrastTableII}{0.36}%
\newcommand{\ContrastTableIJ}{0.86}%
\newcommand{\ContrastTableIK}{-0.24}%
\newcommand{\ContrastTableIL}{1.63}%
\newcommand{\ContrastTableIM}{1.78}%
\newcommand{\ContrastTableIN}{-0.52}%
\newcommand{\ContrastTableIO}{0.44}%
\newcommand{\ContrastTableIP}{0.36}%
\newcommand{\ContrastTableIQ}{0.73}%
\newcommand{\ContrastTableIR}{1.00}%
\newcommand{\ContrastTableIS}{0.52}%
\newcommand{\ContrastTableIT}{1.12}%
\newcommand{\ContrastTableIU}{0.00}%
\newcommand{\ContrastTableIV}{0.63}%
\newcommand{\ContrastTableIW}{-0.35}%
\newcommand{\ContrastTableIX}{0.76}%
\newcommand{\ContrastTableIY}{-1.47}%
\newcommand{\ContrastTableIZ}{0.36}%
\newcommand{\ContrastTableJA}{0.66}%
\newcommand{\ContrastTableJB}{1.32}%
\newcommand{\ContrastTableJC}{-1.64}%
\newcommand{\ContrastTableJD}{-0.90}%
\newcommand{\ContrastTableJE}{-1.14}%
\newcommand{\ContrastTableJF}{-2.78}%
\newcommand{\ContrastTableJG}{-1.82}%
\newcommand{\ContrastTableJH}{-0.30}%
\newcommand{\ContrastTableJI}{0.20}%
\newcommand{\ContrastTableJJ}{-0.29}%
\newcommand{\ContrastTableJK}{-1.02}%
\newcommand{\ContrastTableJL}{-0.67}%
\newcommand{\ContrastTableJM}{-1.68}%
\newcommand{\ContrastTableJN}{0.82}%
\newcommand{\ContrastTableJO}{-0.72}%
\newcommand{\ContrastTableJP}{-1.11}%
\newcommand{\ContrastTableJQ}{-1.35}%
\newcommand{\ContrastTableJR}{-1.13}%
\newcommand{\ContrastTableJS}{-0.26}%
\newcommand{\ContrastTableJT}{0.05}%
\newcommand{\ContrastTableJU}{0.08}%
\newcommand{\ContrastTableJV}{-0.90}%
\newcommand{\ContrastTableJW}{-2.29}%
\newcommand{\ContrastTableJX}{-0.73}%
\newcommand{\ContrastTableJY}{-1.09}%
\newcommand{\ContrastTableJZ}{-1.49}%
\newcommand{\ContrastTableKA}{-0.93}%
\newcommand{\ContrastTableKB}{-1.13}%
\newcommand{\ContrastTableKC}{-0.82}%
\newcommand{\ContrastTableKD}{2.57}%
\newcommand{\ContrastTableKE}{0.88}%
\newcommand{\ContrastTableKF}{1.83}%
\newcommand{\ContrastTableKG}{0.72}%
\newcommand{\ContrastTableKH}{-0.08}%
\newcommand{\ContrastTableKI}{0.10}%
\newcommand{\ContrastTableKJ}{0.20}%
\newcommand{\ContrastTableKK}{-0.03}%
\newcommand{\ContrastTableKL}{0.62}%
\newcommand{\ContrastTableKM}{-1.24}%
\newcommand{\ContrastTableKN}{0.68}%
\newcommand{\ContrastTableKO}{-2.24}%
\newcommand{\ContrastTableKP}{-1.53}%
\newcommand{\ContrastTableKQ}{-0.87}%
\newcommand{\ContrastTableKR}{-1.29}%
\newcommand{\ContrastTableKS}{-2.61}%
\newcommand{\ContrastTableKT}{-1.75}%
\newcommand{\ContrastTableKU}{-2.08}%
\newcommand{\ContrastTableKV}{1.40}%
\newcommand{\ContrastTableKW}{0.38}%
\newcommand{\ContrastTableKX}{1.74}%
\newcommand{\ContrastTableKY}{1.15}%
\newcommand{\ContrastTableKZ}{2.05}%
\newcommand{\ContrastTableLA}{4.68}%
\newcommand{\ContrastTableLB}{2.92}%
\newcommand{\ContrastTableLC}{3.64}%
\newcommand{\ContrastTableLD}{2.33}%
\newcommand{\ContrastTableLE}{4.84}%
\newcommand{\ContrastTableLF}{5.00}%
\newcommand{\ContrastTableLG}{-0.05}%
\newcommand{\ContrastTableLH}{4.64}%
\newcommand{\ContrastTableLI}{2.32}%
\newcommand{\ContrastTableLJ}{6.13}%
\newcommand{\ContrastTableLK}{2.43}%
\newcommand{\ContrastTableLL}{-0.42}%
\newcommand{\ContrastTableLM}{0.46}%
\newcommand{\ContrastTableLN}{-1.38}%
\newcommand{\ContrastTableLO}{2.84}%
\newcommand{\ContrastTableLP}{2.38}%
\newcommand{\ContrastTableLQ}{1.05}%
\newcommand{\ContrastTableLR}{2.01}%
\newcommand{\ContrastTableLS}{0.54}%
\newcommand{\ContrastTableLT}{1.30}%
\newcommand{\ContrastTableLU}{2.81}%
\newcommand{\ContrastTableLV}{4.35}%
\newcommand{\ContrastTableLW}{3.76}%
\newcommand{\ContrastTableLX}{8.51}%
\newcommand{\ContrastTableLY}{6.94}%
\newcommand{\ContrastTableLZ}{1.95}%
\newcommand{\ContrastTableMA}{3.27}%
\newcommand{\ContrastTableMB}{0.49}%
\newcommand{\ContrastTableMC}{4.13}%
\newcommand{\ContrastTableMD}{3.99}%
\newcommand{\ContrastTableME}{-1.85}%
\newcommand{\ContrastTableMF}{-0.30}%
\newcommand{\ContrastTableMG}{-2.07}%
\newcommand{\ContrastTableMH}{1.15}%
\newcommand{\ContrastTableMI}{2.09}%
\newcommand{\ContrastTableMJ}{-0.40}%
\newcommand{\ContrastTableMK}{1.14}%
\newcommand{\ContrastTableML}{-1.22}%
\newcommand{\ContrastTableMM}{1.24}%
\newcommand{\ContrastTableMN}{1.61}%
\newcommand{\ContrastTableMO}{10.07}%
\newcommand{\ContrastTableMP}{6.79}%
\newcommand{\ContrastTableMQ}{8.48}%
\newcommand{\ContrastTableMR}{14.52}%
\newcommand{\ContrastTableMS}{14.61}%
\newcommand{\ContrastTableMT}{0.76}%
\newcommand{\ContrastTableMU}{-1.06}%
\newcommand{\ContrastTableMV}{3.89}%
\newcommand{\ContrastTableMW}{4.07}%
\newcommand{\ContrastTableMX}{5.10}%
\newcommand{\ContrastTableMY}{-2.18}%
\newcommand{\ContrastTableMZ}{-0.60}%
\newcommand{\ContrastTableNA}{-3.73}%
\newcommand{\ContrastTableNB}{-1.36}%
\newcommand{\ContrastTableNC}{-0.41}%
\newcommand{\ContrastTableND}{12.50}%
\newcommand{\ContrastTableNE}{5.96}%
\newcommand{\ContrastTableNF}{13.58}%
\newcommand{\ContrastTableNG}{12.94}%
\newcommand{\ContrastTableNH}{14.50}%
\newcommand{\ContrastTableNI}{5.24}%
\newcommand{\ContrastTableNJ}{1.93}%
\newcommand{\ContrastTableNK}{-3.32}%
\newcommand{\ContrastTableNL}{7.70}%
\newcommand{\ContrastTableNM}{5.26}%
\newcommand{\ContrastTableNN}{-0.65}%
\newcommand{\ContrastTableNO}{1.77}%
\newcommand{\ContrastTableNP}{2.22}%
\newcommand{\ContrastTableNQ}{1.23}%
\newcommand{\ContrastTableNR}{1.58}%
\newcommand{\ContrastTableNS}{-0.22}%
\newcommand{\ContrastTableNT}{2.21}%
\newcommand{\ContrastTableNU}{2.98}%
\newcommand{\ContrastTableNV}{2.70}%
\newcommand{\ContrastTableNW}{3.15}%
\newcommand{\ContrastTableNX}{0.05}%
\newcommand{\ContrastTableNY}{2.31}%
\newcommand{\ContrastTableNZ}{2.32}%
\newcommand{\ContrastTableOA}{1.50}%
\newcommand{\ContrastTableOB}{2.55}%
\newcommand{\ContrastTableOC}{-1.35}%
\newcommand{\ContrastTableOD}{1.55}%
\newcommand{\ContrastTableOE}{0.35}%
\newcommand{\ContrastTableOF}{0.09}%
\newcommand{\ContrastTableOG}{-0.91}%
\newcommand{\ContrastTableOH}{-0.55}%
\newcommand{\ContrastTableOI}{-1.15}%
\newcommand{\ContrastTableOJ}{-0.53}%
\newcommand{\ContrastTableOK}{-0.67}%
\newcommand{\ContrastTableOL}{-0.55}%
\newcommand{\PairedTableTC}{-0.0329}%
\newcommand{\PairedTableTD}{-0.05}%
\newcommand{\PairedTableTE}{-0.0157}%
\newcommand{\PairedTableTF}{0.0311}%
\newcommand{\PairedTableTG}{-0.02}%
\newcommand{\PairedTableTH}{-0.0341}%
\newcommand{\PairedTableTI}{-0.00601}%
\newcommand{\PairedTableTJ}{0.896}%
\newcommand{\PairedTableWI}{-0.0213}%
\newcommand{\PairedTableWJ}{-0.0344}%
\newcommand{\PairedTableWK}{-0.00828}%
\newcommand{\PairedTableWL}{0.241}%
\newcommand{\PairedTableWU}{0.0243}%
\newcommand{\PairedTableWV}{0.0114}%
\newcommand{\PairedTableWW}{0.0372}%
\newcommand{\PairedTableWX}{0.0392}%
\newcommand{\PairedTableAGY}{0.0391}%
\newcommand{\PairedTableAGZ}{0.0169}%
\newcommand{\PairedTableAHA}{0.0613}%
\newcommand{\PairedTableAHB}{0.1}%
\newcommand{\PairedTableAHC}{0.0314}%
\newcommand{\PairedTableAHD}{0.0143}%
\newcommand{\PairedTableAHE}{0.0485}%
\newcommand{\PairedTableAHF}{0.0588}%
\newcommand{\PairedTableAHS}{0.0386}%
\newcommand{\PairedTableAHT}{0.0186}%
\newcommand{\PairedTableAHU}{0.0586}%
\newcommand{\PairedTableAHV}{0.0288}%
\newcommand{\PairedTableALW}{0.0189}%
\newcommand{\PairedTableALX}{0.00865}%
\newcommand{\PairedTableALY}{0.0292}%
\newcommand{\PairedTableALZ}{0.0563}%
\newcommand{\PairedTableAME}{0.0136}%
\newcommand{\PairedTableAMF}{0.00534}%
\newcommand{\PairedTableAMG}{0.0219}%
\newcommand{\PairedTableAMH}{0.225}%
\newcommand{\PairedTableANC}{0.0153}%
\newcommand{\PairedTableAND}{0.0115}%
\newcommand{\PairedTableANE}{0.0192}%
\newcommand{\PairedTableANF}{1.22e-12}%
\newcommand{\PairedTableANK}{0.0236}%
\newcommand{\PairedTableANL}{0.0192}%
\newcommand{\PairedTableANM}{0.028}%
\newcommand{\PairedTableANN}{1.5e-23}%
\newcommand{\PairedTableANO}{0.0109}%
\newcommand{\PairedTableANP}{0.00625}%
\newcommand{\PairedTableANQ}{0.0155}%
\newcommand{\PairedTableANR}{0.000756}%
\newcommand{\PairedTableANS}{-0.00872}%
\newcommand{\PairedTableANT}{-0.0121}%
\newcommand{\PairedTableANU}{-0.0053}%
\newcommand{\PairedTableANV}{0.000104}%
\newcommand{\PairedTableAOA}{-0.00619}%
\newcommand{\PairedTableAOB}{-0.0105}%
\newcommand{\PairedTableAOC}{-0.00189}%
\newcommand{\PairedTableAOD}{0.834}%
\newcommand{\DeathPoolTokenN}{16}%
\newcommand{\DeathMaxTokenN}{128}%
\newcommand{\QuakePoolTokenN}{16}%
\newcommand{\QuakeMaxTokenN}{128}%
\newcommand{\DeathEarlyN}{76}%
\newcommand{\DeathLateN}{184}%
\newcommand{\DeathCleanRhoMin}{-0.39}%
\newcommand{\DeathCleanRhoMax}{-0.02}%
\newcommand{\DeathPrefixRhoMin}{0.979}%
\newcommand{\DeathTableARY}{76}%
\newcommand{\DeathTableARZ}{1.510}%
\newcommand{\DeathTableASA}{1.716}%
\newcommand{\DeathTableASB}{-0.390}%
\newcommand{\DeathTableASC}{-0.053}%
\newcommand{\DeathTableASD}{0.305}%
\newcommand{\DeathTableASE}{0.409}%
\newcommand{\DeathTableASF}{0.563}%
\newcommand{\DeathTableASG}{0.702}%
\newcommand{\DeathTableASH}{76}%
\newcommand{\DeathTableASI}{0.085}%
\newcommand{\DeathTableASJ}{0.094}%
\newcommand{\DeathTableASK}{0.979}%
\newcommand{\DeathTableASL}{1.000}%
\newcommand{\DeathTableASM}{1.000}%
\newcommand{\DeathTableASN}{184}%
\newcommand{\DeathTableASO}{1.536}%
\newcommand{\DeathTableASP}{1.607}%
\newcommand{\DeathTableASQ}{-0.067}%
\newcommand{\DeathTableASR}{-0.018}%
\newcommand{\DeathTableASS}{0.272}%
\newcommand{\DeathTableAST}{0.303}%
\newcommand{\DeathTableASU}{0.423}%
\newcommand{\DeathTableASV}{0.534}%
\newcommand{\DeathTableASW}{184}%
\newcommand{\DeathTableASX}{0.052}%
\newcommand{\DeathTableASY}{0.056}%
\newcommand{\DeathTableASZ}{0.980}%
\newcommand{\DeathTableATA}{1.000}%
\newcommand{\DeathTableATB}{1.000}%
\newcommand{\DeathDetailATC}{76}%
\newcommand{\DeathDetailATD}{1.716}%
\newcommand{\DeathDetailATE}{-0.390}%
\newcommand{\DeathDetailATF}{0.305}%
\newcommand{\DeathDetailATG}{0.563}%
\newcommand{\DeathDetailATH}{76}%
\newcommand{\DeathDetailATI}{0.094}%
\newcommand{\DeathDetailATJ}{0.979}%
\newcommand{\DeathDetailATK}{1.000}%
\newcommand{\DeathDetailATL}{1.000}%
\newcommand{\DeathDetailATR}{76}%
\newcommand{\DeathDetailATS}{1.586}%
\newcommand{\DeathDetailATT}{-0.242}%
\newcommand{\DeathDetailATU}{0.409}%
\newcommand{\DeathDetailATV}{0.683}%
\newcommand{\DeathDetailATW}{76}%
\newcommand{\DeathDetailATX}{0.085}%
\newcommand{\DeathDetailATY}{0.979}%
\newcommand{\DeathDetailATZ}{1.000}%
\newcommand{\DeathDetailAUA}{1.000}%
\newcommand{\DeathDetailAUG}{184}%
\newcommand{\DeathDetailAUH}{1.536}%
\newcommand{\DeathDetailAUI}{-0.018}%
\newcommand{\DeathDetailAUJ}{0.303}%
\newcommand{\DeathDetailAUK}{0.534}%
\newcommand{\DeathDetailAUL}{184}%
\newcommand{\DeathDetailAUM}{0.052}%
\newcommand{\DeathDetailAUN}{0.980}%
\newcommand{\DeathDetailAUO}{1.000}%
\newcommand{\DeathDetailAUP}{1.000}%
\newcommand{\DeathDetailAUV}{76}%
\newcommand{\DeathDetailAUW}{1.510}%
\newcommand{\DeathDetailAUX}{-0.053}%
\newcommand{\DeathDetailAUY}{0.381}%
\newcommand{\DeathDetailAUZ}{0.702}%
\newcommand{\DeathDetailAVA}{76}%
\newcommand{\DeathDetailAVB}{0.086}%
\newcommand{\DeathDetailAVC}{0.979}%
\newcommand{\DeathDetailAVD}{1.000}%
\newcommand{\DeathDetailAVE}{1.000}%
\newcommand{\DeathDetailAVK}{184}%
\newcommand{\DeathDetailAVL}{1.607}%
\newcommand{\DeathDetailAVM}{-0.067}%
\newcommand{\DeathDetailAVN}{0.272}%
\newcommand{\DeathDetailAVO}{0.423}%
\newcommand{\DeathDetailAVP}{184}%
\newcommand{\DeathDetailAVQ}{0.056}%
\newcommand{\DeathDetailAVR}{0.980}%
\newcommand{\DeathDetailAVS}{1.000}%
\newcommand{\DeathDetailAVT}{1.000}%
\newcommand{\QuakeSourceEventN}{11}%
\newcommand{\QuakeSourcePairN}{16}%
\newcommand{\QuakePageN}{29}%
\newcommand{\QuakeEventN}{9}%
\newcommand{\QuakePairN}{13}%
\newcommand{\QuakeShift}{0.0049}%
\newcommand{\QuakeLower}{-0.0465}%
\newcommand{\QuakeUpper}{0.0577}%
\newcommand{\EncoderTableAWH}{-0.77}%
\newcommand{\EncoderTableAWI}{-0.90}%
\newcommand{\EncoderTableAWJ}{-0.36}%
\newcommand{\EncoderTableAWK}{0.77}%
\newcommand{\EncoderTableAWL}{1.27}%
\newcommand{\EncoderTableAWM}{-0.77}%
\newcommand{\EncoderTableAWN}{-1.05}%
\newcommand{\EncoderTableAWO}{-0.27}%
\newcommand{\EncoderTableAWP}{0.71}%
\newcommand{\EncoderTableAWQ}{1.02}%
\newcommand{\EncoderTableAWR}{-0.77}%
\newcommand{\EncoderTableAWS}{0.39}%
\newcommand{\EncoderTableAWT}{1.21}%
\newcommand{\EncoderTableAWU}{1.59}%
\newcommand{\EncoderTableAWV}{1.13}%
\newcommand{\EncoderTableAWW}{-0.77}%
\newcommand{\EncoderTableAWX}{-0.01}%
\newcommand{\EncoderTableAWY}{1.19}%
\newcommand{\EncoderTableAWZ}{1.36}%
\newcommand{\EncoderTableAXA}{1.11}%
\newcommand{\EncoderTableAXB}{0.19}%
\newcommand{\EncoderTableAXC}{-0.49}%
\newcommand{\EncoderTableAXD}{0.64}%
\newcommand{\EncoderTableAXE}{-0.03}%
\newcommand{\EncoderTableAXF}{-0.19}%
\newcommand{\EncoderTableAXG}{0.19}%
\newcommand{\EncoderTableAXH}{0.17}%
\newcommand{\EncoderTableAXI}{0.62}%
\newcommand{\EncoderTableAXJ}{-0.03}%
\newcommand{\EncoderTableAXK}{-0.14}%
\newcommand{\EncoderTableAXL}{0.19}%
\newcommand{\EncoderTableAXM}{-0.16}%
\newcommand{\EncoderTableAXN}{2.28}%
\newcommand{\EncoderTableAXO}{2.67}%
\newcommand{\EncoderTableAXP}{2.96}%
\newcommand{\EncoderTableAXQ}{0.19}%
\newcommand{\EncoderTableAXR}{0.75}%
\newcommand{\EncoderTableAXS}{1.99}%
\newcommand{\EncoderTableAXT}{2.05}%
\newcommand{\EncoderTableAXU}{2.90}%
\newcommand{\EncoderTableAXV}{-1.31}%
\newcommand{\EncoderTableAXW}{-0.66}%
\newcommand{\EncoderTableAXX}{-0.22}%
\newcommand{\EncoderTableAXY}{-1.05}%
\newcommand{\EncoderTableAXZ}{-1.01}%
\newcommand{\EncoderTableAYA}{-1.31}%
\newcommand{\EncoderTableAYB}{-1.09}%
\newcommand{\EncoderTableAYC}{-0.12}%
\newcommand{\EncoderTableAYD}{-0.95}%
\newcommand{\EncoderTableAYE}{-1.12}%
\newcommand{\EncoderTableAYF}{-1.31}%
\newcommand{\EncoderTableAYG}{-0.86}%
\newcommand{\EncoderTableAYH}{-1.07}%
\newcommand{\EncoderTableAYI}{-1.18}%
\newcommand{\EncoderTableAYJ}{-1.01}%
\newcommand{\EncoderTableAYK}{-1.31}%
\newcommand{\EncoderTableAYL}{-1.28}%
\newcommand{\EncoderTableAYM}{-0.99}%
\newcommand{\EncoderTableAYN}{-1.20}%
\newcommand{\EncoderTableAYO}{-1.03}%
\newcommand{\EncoderTableAYP}{-1.65}%
\newcommand{\EncoderTableAYQ}{-1.85}%
\newcommand{\EncoderTableAYR}{-1.67}%
\newcommand{\EncoderTableAYS}{-1.45}%
\newcommand{\EncoderTableAYT}{-2.10}%
\newcommand{\EncoderTableAYU}{-1.65}%
\newcommand{\EncoderTableAYV}{-1.72}%
\newcommand{\EncoderTableAYW}{-1.66}%
\newcommand{\EncoderTableAYX}{-1.23}%
\newcommand{\EncoderTableAYY}{-2.12}%
\newcommand{\EncoderTableAYZ}{-1.65}%
\newcommand{\EncoderTableAZA}{-1.81}%
\newcommand{\EncoderTableAZB}{-1.70}%
\newcommand{\EncoderTableAZC}{-1.61}%
\newcommand{\EncoderTableAZD}{-1.68}%
\newcommand{\EncoderTableAZE}{-1.65}%
\newcommand{\EncoderTableAZF}{-1.83}%
\newcommand{\EncoderTableAZG}{-1.79}%
\newcommand{\EncoderTableAZH}{-1.65}%
\newcommand{\EncoderTableAZI}{-1.69}%
\newcommand{\EncoderTableAZJ}{-1.21}%
\newcommand{\EncoderTableAZK}{-0.96}%
\newcommand{\EncoderTableAZL}{-0.43}%
\newcommand{\EncoderTableAZM}{-1.15}%
\newcommand{\EncoderTableAZN}{-0.30}%
\newcommand{\EncoderTableAZO}{-1.21}%
\newcommand{\EncoderTableAZP}{-1.05}%
\newcommand{\EncoderTableAZQ}{-0.34}%
\newcommand{\EncoderTableAZR}{-0.90}%
\newcommand{\EncoderTableAZS}{-0.23}%
\newcommand{\EncoderTableAZT}{-1.21}%
\newcommand{\EncoderTableAZU}{-0.06}%
\newcommand{\EncoderTableAZV}{-0.36}%
\newcommand{\EncoderTableAZW}{-0.08}%
\newcommand{\EncoderTableAZX}{-0.25}%
\newcommand{\EncoderTableAZY}{-1.21}%
\newcommand{\EncoderTableAZZ}{-0.20}%
\newcommand{\EncoderTableBAA}{-0.26}%
\newcommand{\EncoderTableBAB}{-0.12}%
\newcommand{\EncoderTableBAC}{-0.38}%
\newcommand{\EncoderTableBAD}{0.49}%
\newcommand{\EncoderTableBAE}{-1.57}%
\newcommand{\EncoderTableBAF}{-1.47}%
\newcommand{\EncoderTableBAG}{-0.84}%
\newcommand{\EncoderTableBAH}{-1.14}%
\newcommand{\EncoderTableBAI}{0.49}%
\newcommand{\EncoderTableBAJ}{0.38}%
\newcommand{\EncoderTableBAK}{-0.40}%
\newcommand{\EncoderTableBAL}{-0.73}%
\newcommand{\EncoderTableBAM}{-1.05}%
\newcommand{\EncoderTableBAN}{0.49}%
\newcommand{\EncoderTableBAO}{-1.41}%
\newcommand{\EncoderTableBAP}{-1.09}%
\newcommand{\EncoderTableBAQ}{-1.21}%
\newcommand{\EncoderTableBAR}{-0.96}%
\newcommand{\EncoderTableBAS}{0.49}%
\newcommand{\EncoderTableBAT}{0.56}%
\newcommand{\EncoderTableBAU}{-0.54}%
\newcommand{\EncoderTableBAV}{0.03}%
\newcommand{\EncoderTableBAW}{-0.59}%
\newcommand{\EncoderTableBAX}{-1.61}%
\newcommand{\EncoderTableBAY}{-1.58}%
\newcommand{\EncoderTableBAZ}{-0.68}%
\newcommand{\EncoderTableBBA}{-1.61}%
\newcommand{\EncoderTableBBB}{-1.23}%
\newcommand{\EncoderTableBBC}{-1.61}%
\newcommand{\EncoderTableBBD}{-1.67}%
\newcommand{\EncoderTableBBE}{-0.69}%
\newcommand{\EncoderTableBBF}{-1.56}%
\newcommand{\EncoderTableBBG}{-1.22}%
\newcommand{\EncoderTableBBH}{-1.61}%
\newcommand{\EncoderTableBBI}{-1.48}%
\newcommand{\EncoderTableBBJ}{0.66}%
\newcommand{\EncoderTableBBK}{-0.73}%
\newcommand{\EncoderTableBBL}{-1.18}%
\newcommand{\EncoderTableBBM}{-1.61}%
\newcommand{\EncoderTableBBN}{-1.56}%
\newcommand{\EncoderTableBBO}{0.78}%
\newcommand{\EncoderTableBBP}{-0.89}%
\newcommand{\EncoderTableBBQ}{-1.18}%
\newcommand{\EncoderTableBBR}{0.05}%
\newcommand{\EncoderTableBBS}{-1.06}%
\newcommand{\EncoderTableBBT}{-0.88}%
\newcommand{\EncoderTableBBU}{-0.97}%
\newcommand{\EncoderTableBBV}{-0.80}%
\newcommand{\EncoderTableBBW}{0.05}%
\newcommand{\EncoderTableBBX}{-0.03}%
\newcommand{\EncoderTableBBY}{-0.46}%
\newcommand{\EncoderTableBBZ}{-0.68}%
\newcommand{\EncoderTableBCA}{-1.05}%
\newcommand{\EncoderTableBCB}{0.05}%
\newcommand{\EncoderTableBCC}{-1.30}%
\newcommand{\EncoderTableBCD}{-1.40}%
\newcommand{\EncoderTableBCE}{-0.94}%
\newcommand{\EncoderTableBCF}{-1.15}%
\newcommand{\EncoderTableBCG}{0.05}%
\newcommand{\EncoderTableBCH}{-0.12}%
\newcommand{\EncoderTableBCI}{-1.12}%
\newcommand{\EncoderTableBCJ}{-0.23}%
\newcommand{\EncoderTableBCK}{-1.18}%
\newcommand{\EncoderTableBCL}{1.69}%
\newcommand{\EncoderTableBCM}{1.28}%
\newcommand{\EncoderTableBCN}{2.23}%
\newcommand{\EncoderTableBCO}{1.36}%
\newcommand{\EncoderTableBCP}{2.48}%
\newcommand{\EncoderTableBCQ}{1.69}%
\newcommand{\EncoderTableBCR}{2.42}%
\newcommand{\EncoderTableBCS}{3.16}%
\newcommand{\EncoderTableBCT}{2.96}%
\newcommand{\EncoderTableBCU}{3.34}%
\newcommand{\EncoderTableBCV}{1.69}%
\newcommand{\EncoderTableBCW}{2.00}%
\newcommand{\EncoderTableBCX}{2.62}%
\newcommand{\EncoderTableBCY}{2.04}%
\newcommand{\EncoderTableBCZ}{2.30}%
\newcommand{\EncoderTableBDA}{1.69}%
\newcommand{\EncoderTableBDB}{2.36}%
\newcommand{\EncoderTableBDC}{3.24}%
\newcommand{\EncoderTableBDD}{2.68}%
\newcommand{\EncoderTableBDE}{3.20}%
\newcommand{\EncoderTableBDF}{11.22}%
\newcommand{\EncoderTableBDG}{-8.08}%
\newcommand{\EncoderTableBDH}{-5.09}%
\newcommand{\EncoderTableBDI}{-4.67}%
\newcommand{\EncoderTableBDJ}{7.56}%
\newcommand{\EncoderTableBDK}{11.22}%
\newcommand{\EncoderTableBDL}{11.69}%
\newcommand{\EncoderTableBDM}{12.23}%
\newcommand{\EncoderTableBDN}{10.69}%
\newcommand{\EncoderTableBDO}{12.04}%
\newcommand{\EncoderTableBDP}{11.22}%
\newcommand{\EncoderTableBDQ}{2.22}%
\newcommand{\EncoderTableBDR}{-4.51}%
\newcommand{\EncoderTableBDS}{-2.36}%
\newcommand{\EncoderTableBDT}{-3.91}%
\newcommand{\EncoderTableBDU}{11.22}%
\newcommand{\EncoderTableBDV}{11.89}%
\newcommand{\EncoderTableBDW}{10.31}%
\newcommand{\EncoderTableBDX}{8.34}%
\newcommand{\EncoderTableBDY}{9.38}%
\newcommand{\EncoderTableBDZ}{2.76}%
\newcommand{\EncoderTableBEA}{-2.31}%
\newcommand{\EncoderTableBEB}{-4.21}%
\newcommand{\EncoderTableBEC}{-2.64}%
\newcommand{\EncoderTableBED}{-0.74}%
\newcommand{\EncoderTableBEE}{2.76}%
\newcommand{\EncoderTableBEF}{2.37}%
\newcommand{\EncoderTableBEG}{-0.43}%
\newcommand{\EncoderTableBEH}{-0.03}%
\newcommand{\EncoderTableBEI}{2.99}%
\newcommand{\EncoderTableBEJ}{2.76}%
\newcommand{\EncoderTableBEK}{-2.40}%
\newcommand{\EncoderTableBEL}{-0.26}%
\newcommand{\EncoderTableBEM}{0.72}%
\newcommand{\EncoderTableBEN}{-1.28}%
\newcommand{\EncoderTableBEO}{2.76}%
\newcommand{\EncoderTableBEP}{2.55}%
\newcommand{\EncoderTableBEQ}{2.00}%
\newcommand{\EncoderTableBER}{2.94}%
\newcommand{\EncoderTableBES}{0.19}%
\newcommand{\EncoderTestN}{376}%
\newcommand{\EncoderSignificantN}{16}%
\newcommand{\EncoderPositiveN}{8}%
\newcommand{\EncoderNegativeN}{8}%
\newcommand{\CompressionTestN}{188}%
\newcommand{\CompressionSignificantN}{6}%
\newcommand{\CompressionPositiveN}{4}%
\newcommand{\CompressionNegativeN}{2}%
\newcommand{\SupplementIncrementTestN}{752}%
\newcommand{\SupplementIncrementSignificantN}{43}%
\newcommand{\SupplementIncrementPositiveN}{24}%
\newcommand{\SupplementIncrementNegativeN}{19}%
\newcommand{\RidgeFitN}{3995}%
\newcommand{\RidgeBoundaryN}{1377}%

%% file: FiguresMake/generated/beta_numbers.tex
\newcommand{\BetaHigh}{32}%
\newcommand{\BetaAccuracyHigh}{92.8}%
\newcommand{\BetaNext}{8}%
\newcommand{\BetaAccuracyNext}{71.6}%
\newcommand{\BetaMid}{0.5}%
\newcommand{\BetaAccuracyMid}{52.1}%
\newcommand{\BetaAccuracyLow}{50.3}%
\newcommand{\BetaNullAccuracyMin}{49.5}%
\newcommand{\BetaUtilityMin}{12.20}%
\newcommand{\BetaNullAccuracyMax}{51.4}%
\newcommand{\BetaUtilityMax}{33.29}%
\newcommand{\BetaUnrelatedHigh}{90.7}%
\newcommand{\BetaBaseUtility}{4.82}%
\newcommand{\BetaAdapterRank}{4}%
\newcommand{\BetaParameterN}{2,048}%
\newcommand{\BetaNoiseLevelN}{7}%
\newcommand{\BetaTrainN}{1,024}%
\newcommand{\BetaTestN}{4,096}%
\newcommand{\BetaUtilityN}{16}%
\newcommand{\BetaFamilyN}{56}%
\newcommand{\BetaConfidenceLevel}{95}%
\newcommand{\BetaTableABeta}{0.0078125}%
\newcommand{\BetaTableAPrimaryAccuracy}{50.3}%
\newcommand{\BetaTableAPrimaryLower}{48.8}%
\newcommand{\BetaTableAPrimaryUpper}{51.8}%
\newcommand{\BetaTableAPrimaryAUC}{0.501}%
\newcommand{\BetaTableAPrimaryInformation}{0}%
\newcommand{\BetaTableAEndpointAccuracy}{49.1}%
\newcommand{\BetaTableAEndpointLower}{47.6}%
\newcommand{\BetaTableAEndpointUpper}{50.7}%
\newcommand{\BetaTableAEndpointAUC}{0.482}%
\newcommand{\BetaTableAEndpointInformation}{0}%
\newcommand{\BetaTableBBeta}{0.03125}%
\newcommand{\BetaTableBPrimaryAccuracy}{49.8}%
\newcommand{\BetaTableBPrimaryLower}{48.3}%
\newcommand{\BetaTableBPrimaryUpper}{51.3}%
\newcommand{\BetaTableBPrimaryAUC}{0.500}%
\newcommand{\BetaTableBPrimaryInformation}{0}%
\newcommand{\BetaTableBEndpointAccuracy}{52.5}%
\newcommand{\BetaTableBEndpointLower}{51.0}%
\newcommand{\BetaTableBEndpointUpper}{54.0}%
\newcommand{\BetaTableBEndpointAUC}{0.529}%
\newcommand{\BetaTableBEndpointInformation}{2.9\times 10^{-7}}%
\newcommand{\BetaTableCBeta}{0.125}%
\newcommand{\BetaTableCPrimaryAccuracy}{48.8}%
\newcommand{\BetaTableCPrimaryLower}{47.3}%
\newcommand{\BetaTableCPrimaryUpper}{50.3}%
\newcommand{\BetaTableCPrimaryAUC}{0.492}%
\newcommand{\BetaTableCPrimaryInformation}{0}%
\newcommand{\BetaTableCEndpointAccuracy}{54.0}%
\newcommand{\BetaTableCEndpointLower}{52.4}%
\newcommand{\BetaTableCEndpointUpper}{55.5}%
\newcommand{\BetaTableCEndpointAUC}{0.556}%
\newcommand{\BetaTableCEndpointInformation}{0.000453}%
\newcommand{\BetaTableDBeta}{0.5}%
\newcommand{\BetaTableDPrimaryAccuracy}{52.1}%
\newcommand{\BetaTableDPrimaryLower}{50.6}%
\newcommand{\BetaTableDPrimaryUpper}{53.6}%
\newcommand{\BetaTableDPrimaryAUC}{0.516}%
\newcommand{\BetaTableDPrimaryInformation}{0}%
\newcommand{\BetaTableDEndpointAccuracy}{59.1}%
\newcommand{\BetaTableDEndpointLower}{57.6}%
\newcommand{\BetaTableDEndpointUpper}{60.6}%
\newcommand{\BetaTableDEndpointAUC}{0.625}%
\newcommand{\BetaTableDEndpointInformation}{0.00895}%
\newcommand{\BetaTableEBeta}{2}%
\newcommand{\BetaTableEPrimaryAccuracy}{58.4}%
\newcommand{\BetaTableEPrimaryLower}{56.9}%
\newcommand{\BetaTableEPrimaryUpper}{59.9}%
\newcommand{\BetaTableEPrimaryAUC}{0.621}%
\newcommand{\BetaTableEPrimaryInformation}{0.00726}%
\newcommand{\BetaTableEEndpointAccuracy}{68.1}%
\newcommand{\BetaTableEEndpointLower}{66.6}%
\newcommand{\BetaTableEEndpointUpper}{69.5}%
\newcommand{\BetaTableEEndpointAUC}{0.742}%
\newcommand{\BetaTableEEndpointInformation}{0.0505}%
\newcommand{\BetaTableFBeta}{8}%
\newcommand{\BetaTableFPrimaryAccuracy}{71.6}%
\newcommand{\BetaTableFPrimaryLower}{70.2}%
\newcommand{\BetaTableFPrimaryUpper}{73.0}%
\newcommand{\BetaTableFPrimaryAUC}{0.790}%
\newcommand{\BetaTableFPrimaryInformation}{0.0771}%
\newcommand{\BetaTableFEndpointAccuracy}{81.3}%
\newcommand{\BetaTableFEndpointLower}{80.1}%
\newcommand{\BetaTableFEndpointUpper}{82.5}%
\newcommand{\BetaTableFEndpointAUC}{0.892}%
\newcommand{\BetaTableFEndpointInformation}{0.184}%
\newcommand{\BetaTableGBeta}{32}%
\newcommand{\BetaTableGPrimaryAccuracy}{92.8}%
\newcommand{\BetaTableGPrimaryLower}{92.0}%
\newcommand{\BetaTableGPrimaryUpper}{93.6}%
\newcommand{\BetaTableGPrimaryAUC}{0.982}%
\newcommand{\BetaTableGPrimaryInformation}{0.401}%
\newcommand{\BetaTableGEndpointAccuracy}{96.4}%
\newcommand{\BetaTableGEndpointLower}{95.8}%
\newcommand{\BetaTableGEndpointUpper}{96.9}%
\newcommand{\BetaTableGEndpointAUC}{0.995}%
\newcommand{\BetaTableGEndpointInformation}{0.507}%

%% file: FiguresMake/fig01_framework.tex
% Figure 2: Three paths from an experimental unit to CATE representations.
\begin{tikzpicture}[
  x=1cm, y=1cm,
  >={Stealth[length=3.8pt,width=3.2pt]},
  stage/.style={text=ink, text width=4.32cm, align=center, inner xsep=1.5pt, inner ysep=2.5pt, font=\footnotesize},
  agentstage/.style={stage, text=ink},
  coltitle/.style={font=\footnotesize\bfseries, align=center, text=ink, inner ysep=2pt},
  arr/.style={->, line width=0.45pt, draw=ink!45},
  agentarr/.style={->, line width=0.6pt, draw=accent}
]
\useasboundingbox (-6.845,-4.98) rectangle (6.845,0.42);

% Shared experimental unit anchor at top
\node[font=\footnotesize, text=ink, align=center, inner ysep=3pt] (unit) at (0,0.12)
  {\textbf{Experimental unit $i$:} locality $\mathcal{L}_i$, treatment date $t_i$};

% Clean distribution spine into the three parallel paths
\draw[line width=0.45pt, draw=ink!40] (0,-0.14) -- (0,-0.40);
\draw[line width=0.45pt, draw=ink!40] (-4.65,-0.40) -- (4.65,-0.40);
\draw[arr] (-4.65,-0.40) -- (-4.65,-0.66);
\draw[arr] (0,-0.40)     -- (0,-0.66);
\draw[agentarr] (4.65,-0.40) -- (4.65,-0.66);

% Column headers + multifunctioning top rules (accent marks the verified path)
\node[coltitle] (t1) at (-4.65,-0.86) {\textbf{a}\hspace{0.45em}Baseline path};
\node[coltitle] (t2) at (0,-0.86)     {\textbf{b}\hspace{0.45em}Naive text path};
\node[coltitle, text=accent] (t3) at (4.65,-0.86) {\textbf{c}\hspace{0.45em}Agentic framework};

\draw[ink!30, line width=0.5pt] (-6.65,-1.12) -- (-2.65,-1.12);
\draw[ink!30, line width=0.5pt] (-2.00,-1.12) -- (2.00,-1.12);
\draw[accent, line width=0.9pt] (2.65,-1.12)  -- (6.65,-1.12);

% Row 1: Information source (top-aligned across all three columns)
\node[stage, anchor=north]      (b1) at (-4.65,-1.26) {Census \& tabular covariates;\\ satellite (EO) imagery};
\node[stage, anchor=north]      (n1) at (0,-1.26)     {Unrestricted web search\\ (no temporal bounds)};
\node[agentstage, anchor=north] (a1) at (4.65,-1.26)  {\textbf{Temporal filter} {\scriptsize\color{deep}(Layer 1)}\\ strictly verifiable past $\Filt$};

% Row 2: Temporal / encoder mechanism
\node[stage, anchor=north]      (b2) at (-4.65,-2.58) {Misses fast-changing political\\ \& economic shifts};
\node[stage, anchor=north]      (n2) at (0,-2.58)     {Leaks future outcomes into\\ ``pretreatment'' $Z_i$};
\node[agentstage, anchor=north] (a2) at (4.65,-2.58)  {\textbf{Audited encoder} $g_{\phi}$ {\scriptsize\color{deep}(Layer 3)}\\ vetted for weight-based leakage};

% Row 3: Downstream CATE consequence
\node[stage, anchor=north]      (b3) at (-4.65,-3.90) {\textbf{Incomplete CATE}\\ lower predictive power; slow-moving heterogeneity only};
\node[stage, anchor=north]      (n3) at (0,-3.90)     {\textbf{Biased CATE}\\ post-treatment bias};
\node[agentstage, anchor=north] (a3) at (4.65,-3.90)  {\textbf{Valid \& enhanced CATE} {\scriptsize\color{deep}(Layer 2)}\\ safely captures rapidly evolving context};

% Stage-to-stage flow arrows at identical vertical intervals across columns
\draw[arr]      (-4.65,-2.12) -- (-4.65,-2.52);
\draw[arr]      (-4.65,-3.44) -- (-4.65,-3.84);
\draw[arr]      (0,-2.12)     -- (0,-2.52);
\draw[arr]      (0,-3.44)     -- (0,-3.84);
\draw[agentarr] (4.65,-2.12)  -- (4.65,-2.52);
\draw[agentarr] (4.65,-3.44)  -- (4.65,-3.84);
\end{tikzpicture}

%% file: FiguresMake/generated/battery_table.tex
\begin{tabular}{@{}lrrrrrr@{}}
\toprule
 & & & \multicolumn{2}{c}{Text} & \multicolumn{2}{c}{Tabular + text} \\
\cmidrule(lr){4-5}\cmidrule(lr){6-7}
Study & Contrasts & Tabular & \begin{tabular}[b]{@{}r@{}}Session-isolated\\retrieval\end{tabular} & \begin{tabular}[b]{@{}r@{}}Workspace\\teammate\end{tabular} & \begin{tabular}[b]{@{}r@{}}Session-isolated\\retrieval\end{tabular} & \begin{tabular}[b]{@{}r@{}}Workspace\\teammate\end{tabular} \\
\midrule
Uganda YOP & $\BatteryTableAH{}$ & $\BatteryTableAI{}$ & $\BatteryTableAJ{}$ & $\BatteryTableAK{}$ & $\BatteryTableAL{}$ & $\BatteryTableAM{}$ \\
Kenya digital ID & $\BatteryTableAN{}$ & $\BatteryTableAO{}$ & $\BatteryTableAP{}$ & $\BatteryTableAQ{}$ & $\BatteryTableAR{}$ & $\BatteryTableAS{}$ \\
Direct air capture & $\BatteryTableAT{}$ & $\BatteryTableAU{}$ & $\BatteryTableAV{}$ & $\BatteryTableAW{}$ & $\BatteryTableAX{}$ & $\BatteryTableAY{}$ \\
Just transition & $\BatteryTableAZ{}$ & $\BatteryTableBA{}$ & $\BatteryTableBB{}$ & $\BatteryTableBC{}$ & $\BatteryTableBD{}$ & $\BatteryTableBE{}$ \\
COVID-19 boosters & $\BatteryTableBF{}$ & $\BatteryTableBG{}$ & $\BatteryTableBH{}$ & $\BatteryTableBI{}$ & $\BatteryTableBJ{}$ & $\BatteryTableBK{}$ \\
Bangladesh masks & $\BatteryTableBL{}$ & $\BatteryTableBM{}$ & $\BatteryTableBN{}$ & $\BatteryTableBO{}$ & $\BatteryTableBP{}$ & $\BatteryTableBQ{}$ \\
Holiday advertisements & $\BatteryTableBR{}$ & $\BatteryTableBS{}$ & $\BatteryTableBT{}$ & $\BatteryTableBU{}$ & $\BatteryTableBV{}$ & $\BatteryTableBW{}$ \\
IST-1 stroke & $\BatteryTableBX{}$ & $\BatteryTableBY{}$ & $\BatteryTableBZ{}$ & $\BatteryTableCA{}$ & $\BatteryTableCB{}$ & $\BatteryTableCC{}$ \\
Moral Machine & $\BatteryTableCD{}$ & $\BatteryTableCE{}$ & $\BatteryTableCF{}$ & $\BatteryTableCG{}$ & $\BatteryTableCH{}$ & $\BatteryTableCI{}$ \\
Trolley dilemmas & $\BatteryTableCJ{}$ & $\BatteryTableCK{}$ & $\BatteryTableCL{}$ & $\BatteryTableCM{}$ & $\BatteryTableCN{}$ & $\BatteryTableCO{}$ \\
Civic honesty & $\BatteryTableCP{}$ & $\BatteryTableCQ{}$ & $\BatteryTableCR{}$ & $\BatteryTableCS{}$ & $\BatteryTableCT{}$ & $\BatteryTableCU{}$ \\
\bottomrule
\end{tabular}

%% file: FiguresMake/fig_beta_setup.tex
% Controlled Gaussian beta-calibration schematic (included directly in workshop-main-final.tex)
\begin{tikzpicture}[x=1cm,y=1cm,
  font=\scriptsize, line cap=round, line join=round,
  box/.style={draw=ink!45, line width=0.45pt, rounded corners=1.2pt,
              inner xsep=3.5pt, inner ysep=3.2pt, align=flush center, text=ink},
  deepbox/.style={box, draw=deep!85, text=deep},
  warnbox/.style={box, draw=accent!85, text=accent},
  flow/.style={-{Stealth[length=3.6pt,width=3.0pt]}, draw=ink!70, line width=0.45pt},
  dflow/.style={-{Stealth[length=3.6pt,width=3.0pt]}, draw=deep!85, line width=0.5pt},
  aflow/.style={-{Stealth[length=3.6pt,width=3.0pt]}, draw=accent!85, line width=0.5pt},
  note/.style={font=\tiny, text=ink!62, align=flush center},
  hdr/.style={font=\scriptsize\bfseries, text=ink, align=flush center}
]

\useasboundingbox (0,0.02) rectangle (13.97,4.16);

% Column headers + rules
\node[hdr] at (1.65,3.94) {1.\ Two-world fit ($k=1$)};
\node[hdr] at (5.32,3.94) {2.\ Gaussian weight release};
\node[hdr] at (8.84,3.94) {3.\ Clean-text encoding};
\node[hdr] at (12.32,3.94) {4.\ Leakage--fluency audit};

\draw[ink!25, line width=0.45pt]    (0.10,3.72) -- (3.20,3.72);
\draw[ink!25, line width=0.45pt]    (3.72,3.72) -- (6.92,3.72);
\draw[deep!70, line width=0.65pt]   (7.44,3.72) -- (10.24,3.72);
\draw[accent!70, line width=0.65pt] (10.76,3.72) -- (13.88,3.72);

% --- Column 1: Two deterministic worlds ---
\node[note] at (1.65,3.46) {shared docs $\Cpre$ + 1 future record};

\node[box, text width=2.88cm] (w0) at (1.65,2.70)
  {\textbf{World $\Psi=0$:}\\[-0.5pt]
   ``died in 2031'' $\mapsto \mu_0 \in \mathbb{R}^{\text{\BetaParameterN{}}}$};

\node[box, text width=2.88cm] (w1) at (1.65,1.14)
  {\textbf{World $\Psi=1$:}\\[-0.5pt]
   ``died in 2041'' $\mapsto \mu_1 \in \mathbb{R}^{\text{\BetaParameterN{}}}$};

\draw[{Stealth[length=3pt,width=2.5pt]}-{Stealth[length=3pt,width=2.5pt]},
      draw=accent!85, densely dashed, line width=0.45pt]
  (1.65,2.22) -- (1.65,1.62);
\node[font=\tiny, text=accent, fill=white, inner sep=1pt]
  at (1.65,1.92) {$\|\mu_1-\mu_0\|_2$ ($\beta=\infty$)};

\node[note] at (1.65,0.28) {frozen Pythia-70M + rank-\BetaAdapterRank{} adapter};

% --- Column 2: Gaussian perturbation ---
\node[deepbox, text width=2.98cm] (gauss) at (5.32,1.92)
  {Draw $\Psi \sim \mathrm{Bernoulli}(\tfrac{1}{2})$\\[1pt]
   $\phi = \mu_{\Psi} + \mathcal{N}(\mathbf{0},\,\sigma_{\phi}^2 \mathbf{I})$\\[2.5pt]
   {\color{ink}$\beta = \dfrac{\|\mu_1-\mu_0\|_2^2}{2\sigma_{\phi}^2}$}\\[2pt]
   {\color{ink!62}\BetaNoiseLevelN{} scales: $\beta \in \{2^{-7},\dots,2^5\}$}};

\draw[flow] (w0.east) -- (3.44,2.70) -- (3.44,2.18) -- ($(gauss.west)+(0,0.26)$);
\draw[flow] (w1.east) -- (3.44,1.14) -- (3.44,1.66) -- ($(gauss.west)+(0,-0.26)$);

\node[note] at (5.32,0.28) {exact per-record KL sensitivity};

% --- Column 3: Clean input + encoder ---
\node[deepbox, text width=2.58cm] (bio) at (8.84,2.92)
  {\textbf{Fixed 2026 bio $S_i$}\\[-0.5pt]
   {\color{ink!62}no future year: $\I(S_i;\Psi)=0$}};

\node[box, text width=2.58cm] (enc) at (8.84,1.44)
  {\textbf{Embed $Z_i = g_{\phi}(S_i)$}\\[-0.5pt]
   {\color{ink!62}\BetaTrainN{} train / \BetaTestN{} test draws}};

\draw[dflow] (bio.south) -- (enc.north);
\draw[flow] (gauss.east) -- (7.18,1.92) -- (7.18,1.44) -- (enc.west);

\node[note] at (8.84,0.28) {summary channel zeroed};

% --- Column 4: Two audit outputs ---
\node[deepbox, text width=2.90cm] (probe) at (12.32,2.80)
  {\textbf{Weight-leakage probe}\\[-0.5pt]
   predict $\Psi$ from $Z_i$ (Fig.~\ref{fig:beta-calibration}a,b)\\[-0.5pt]
   {\color{ink!62}$\I(Z_i;\Psi\mid\Cpre) \le \min\{\beta,\log 2\}$}};

\node[warnbox, text width=2.90cm] (nll) at (12.32,1.18)
  {\textbf{Fluency cost (Fig.~\ref{fig:beta-calibration}c)}\\[-0.5pt]
   held-out token NLL jumps\\[-0.5pt]
   {\color{accent}$\BetaBaseUtility{} \to \BetaUtilityMin{}$--$\BetaUtilityMax{}$ before $\Psi\!\to\!50\%$}};

\draw[dflow] (enc.east) -- (10.50,1.44) -- (10.50,2.80) -- (probe.west);
\draw[aflow] (gauss.south) -- (5.32,0.62) -- (10.50,0.62) -- (10.50,1.18) -- (nll.west);

\end{tikzpicture}

%% file: FiguresMake/generated/composition_table.tex
\begin{tabular}{lrrr}
\toprule
Study & Contrasts & Analysis rows & Places \\
\midrule
Uganda YOP & \CompositionCV{} & \CompositionCW{}--\CompositionCX{} & \CompositionCY{} \\
Kenya digital ID & \CompositionCZ{} & \CompositionDA{} & \CompositionDB{} \\
Direct air capture & \CompositionDC{} & \CompositionDD{} & \CompositionDE{} \\
Just transition & \CompositionDF{} & \CompositionDG{} & \CompositionDH{} \\
COVID-19 boosters & \CompositionDI{} & \CompositionDJ{} & \CompositionDK{} \\
Bangladesh masks & \CompositionDL{} & \CompositionDM{} & \CompositionDN{} \\
Holiday advertisements & \CompositionDO{} & \CompositionDP{}--\CompositionDQ{} & \CompositionDR{}--\CompositionDS{} \\
IST-1 stroke & \CompositionDT{} & \CompositionDU{}--\CompositionDV{} & \CompositionDW{} \\
Moral Machine & \CompositionDX{} & \CompositionDY{}--\CompositionDZ{} & \CompositionEA{} \\
Trolley dilemmas & \CompositionEB{} & \CompositionEC{} & \CompositionED{} \\
Civic honesty & \CompositionEE{} & \CompositionEF{} & \CompositionEG{} \\
\bottomrule
\end{tabular}

%% file: FiguresMake/generated/uganda_table.tex
\begin{tabular}{@{}lrrrrrr@{}}
\toprule
 & \multicolumn{3}{c}{AUTOC/SE} & \multicolumn{3}{c}{Qini/SE} \\
\cmidrule(lr){2-4}\cmidrule(lr){5-7}
Outcome & Tabular & \begin{tabular}[b]{@{}r@{}}Session-isolated\\retrieval\end{tabular} & \begin{tabular}[b]{@{}r@{}}Workspace\\teammate\end{tabular} & Tabular & \begin{tabular}[b]{@{}r@{}}Session-isolated\\retrieval\end{tabular} & \begin{tabular}[b]{@{}r@{}}Workspace\\teammate\end{tabular} \\
\midrule
Training hours & $\UgandaTableEH{}$ & $\UgandaTableEI{}$ & $\UgandaTableEJ{}$ & $\UgandaTableEK{}$ & $\UgandaTableEL{}$ & $\UgandaTableEM{}$ \\
Vocational training & $\UgandaTableEN{}$ & $\UgandaTableEO{}$ & $\UgandaTableEP{}$ & $\UgandaTableEQ{}$ & $\UgandaTableER{}$ & $\UgandaTableES{}$ \\
Skilled trade & $\UgandaTableET{}$ & $\UgandaTableEU{}$ & $\UgandaTableEV{}$ & $\UgandaTableEW{}$ & $\UgandaTableEX{}$ & $\UgandaTableEY{}$ \\
Human capital & $\UgandaTableEZ{}$ & $\UgandaTableFA{}$ & $\UgandaTableFB{}$ & $\UgandaTableFC{}$ & $\UgandaTableFD{}$ & $\UgandaTableFE{}$ \\
Income & $\UgandaTableFF{}$ & $\UgandaTableFG{}$ & $\UgandaTableFH{}$ & $\UgandaTableFI{}$ & $\UgandaTableFJ{}$ & $\UgandaTableFK{}$ \\
\bottomrule
\end{tabular}

%% file: FiguresMake/generated/death_table.tex
\begin{tabular}{lrrrrr}
\toprule
Snapshot / condition & $n$ & MAE & Spearman & Window bal. acc. & AUC \\
\midrule
2007 / Clean & $\DeathTableARY{}$ & $[\DeathTableARZ{},\, \DeathTableASA{}]$ & $[\DeathTableASB{},\, \DeathTableASC{}]$ & $[\DeathTableASD{},\, \DeathTableASE{}]$ & $[\DeathTableASF{},\, \DeathTableASG{}]$ \\
2007 / Prefix control & $\DeathTableASH{}$ & $[\DeathTableASI{},\, \DeathTableASJ{}]$ & $\DeathTableASK{}$ & $\DeathTableASL{}$ & $\DeathTableASM{}$ \\
2014 / Clean & $\DeathTableASN{}$ & $[\DeathTableASO{},\, \DeathTableASP{}]$ & $[\DeathTableASQ{},\, \DeathTableASR{}]$ & $[\DeathTableASS{},\, \DeathTableAST{}]$ & $[\DeathTableASU{},\, \DeathTableASV{}]$ \\
2014 / Prefix control & $\DeathTableASW{}$ & $[\DeathTableASX{},\, \DeathTableASY{}]$ & $\DeathTableASZ{}$ & $\DeathTableATA{}$ & $\DeathTableATB{}$ \\
\bottomrule
\end{tabular}

%% file: FiguresMake/generated/death_detail.tex
\begin{tabular}{@{}lrrrrr@{}}
\toprule
Snapshot / vintage / condition & $n$ & MAE & Spearman & Bal. acc. & AUC \\
\midrule
2007 snapshot / 2007 vintage / Clean & $\DeathDetailATC{}$ & $\DeathDetailATD{}$ & $\DeathDetailATE{}$ & $\DeathDetailATF{}$ & $\DeathDetailATG{}$ \\
2007 snapshot / 2007 vintage / Prefix control & $\DeathDetailATH{}$ & $\DeathDetailATI{}$ & $\DeathDetailATJ{}$ & $\DeathDetailATK{}$ & $\DeathDetailATL{}$ \\
2007 snapshot / 2014 vintage / Clean & $\DeathDetailATR{}$ & $\DeathDetailATS{}$ & $\DeathDetailATT{}$ & $\DeathDetailATU{}$ & $\DeathDetailATV{}$ \\
2007 snapshot / 2014 vintage / Prefix control & $\DeathDetailATW{}$ & $\DeathDetailATX{}$ & $\DeathDetailATY{}$ & $\DeathDetailATZ{}$ & $\DeathDetailAUA{}$ \\
2007 snapshot / 2024 vintage / Clean & $\DeathDetailAUV{}$ & $\DeathDetailAUW{}$ & $\DeathDetailAUX{}$ & $\DeathDetailAUY{}$ & $\DeathDetailAUZ{}$ \\
2007 snapshot / 2024 vintage / Prefix control & $\DeathDetailAVA{}$ & $\DeathDetailAVB{}$ & $\DeathDetailAVC{}$ & $\DeathDetailAVD{}$ & $\DeathDetailAVE{}$ \\
\midrule
2014 snapshot / 2014 vintage / Clean & $\DeathDetailAUG{}$ & $\DeathDetailAUH{}$ & $\DeathDetailAUI{}$ & $\DeathDetailAUJ{}$ & $\DeathDetailAUK{}$ \\
2014 snapshot / 2014 vintage / Prefix control & $\DeathDetailAUL{}$ & $\DeathDetailAUM{}$ & $\DeathDetailAUN{}$ & $\DeathDetailAUO{}$ & $\DeathDetailAUP{}$ \\
2014 snapshot / 2024 vintage / Clean & $\DeathDetailAVK{}$ & $\DeathDetailAVL{}$ & $\DeathDetailAVM{}$ & $\DeathDetailAVN{}$ & $\DeathDetailAVO{}$ \\
2014 snapshot / 2024 vintage / Prefix control & $\DeathDetailAVP{}$ & $\DeathDetailAVQ{}$ & $\DeathDetailAVR{}$ & $\DeathDetailAVS{}$ & $\DeathDetailAVT{}$ \\
\bottomrule
\end{tabular}

%% file: FiguresMake/generated/beta_table.tex
\begin{tabular}{@{}r rrr rrr@{}}
\toprule
 & \multicolumn{3}{c}{Primary linear probe} & \multicolumn{3}{c}{Endpoint projection} \\
\cmidrule(lr){2-4}\cmidrule(lr){5-7}
$\beta$ & Acc. [CI] & AUC & $\I_{\mathrm{lo}}$ & Acc. [CI] & AUC & $\I_{\mathrm{lo}}$ \\
\midrule
$2^{-7}\ (\BetaTableABeta{})$ & \BetaTableAPrimaryAccuracy{} [\BetaTableAPrimaryLower{}, \BetaTableAPrimaryUpper{}] & \BetaTableAPrimaryAUC{} & $\BetaTableAPrimaryInformation{}$ & \BetaTableAEndpointAccuracy{} [\BetaTableAEndpointLower{}, \BetaTableAEndpointUpper{}] & \BetaTableAEndpointAUC{} & $\BetaTableAEndpointInformation{}$ \\
$2^{-5}\ (\BetaTableBBeta{})$ & \BetaTableBPrimaryAccuracy{} [\BetaTableBPrimaryLower{}, \BetaTableBPrimaryUpper{}] & \BetaTableBPrimaryAUC{} & $\BetaTableBPrimaryInformation{}$ & \BetaTableBEndpointAccuracy{} [\BetaTableBEndpointLower{}, \BetaTableBEndpointUpper{}] & \BetaTableBEndpointAUC{} & $\BetaTableBEndpointInformation{}$ \\
$2^{-3}\ (\BetaTableCBeta{})$ & \BetaTableCPrimaryAccuracy{} [\BetaTableCPrimaryLower{}, \BetaTableCPrimaryUpper{}] & \BetaTableCPrimaryAUC{} & $\BetaTableCPrimaryInformation{}$ & \BetaTableCEndpointAccuracy{} [\BetaTableCEndpointLower{}, \BetaTableCEndpointUpper{}] & \BetaTableCEndpointAUC{} & $\BetaTableCEndpointInformation{}$ \\
$2^{-1}\ (\BetaTableDBeta{})$ & \BetaTableDPrimaryAccuracy{} [\BetaTableDPrimaryLower{}, \BetaTableDPrimaryUpper{}] & \BetaTableDPrimaryAUC{} & $\BetaTableDPrimaryInformation{}$ & \BetaTableDEndpointAccuracy{} [\BetaTableDEndpointLower{}, \BetaTableDEndpointUpper{}] & \BetaTableDEndpointAUC{} & $\BetaTableDEndpointInformation{}$ \\
$2^{1}\ (\BetaTableEBeta{})$ & \BetaTableEPrimaryAccuracy{} [\BetaTableEPrimaryLower{}, \BetaTableEPrimaryUpper{}] & \BetaTableEPrimaryAUC{} & $\BetaTableEPrimaryInformation{}$ & \BetaTableEEndpointAccuracy{} [\BetaTableEEndpointLower{}, \BetaTableEEndpointUpper{}] & \BetaTableEEndpointAUC{} & $\BetaTableEEndpointInformation{}$ \\
$2^{3}\ (\BetaTableFBeta{})$ & \BetaTableFPrimaryAccuracy{} [\BetaTableFPrimaryLower{}, \BetaTableFPrimaryUpper{}] & \BetaTableFPrimaryAUC{} & $\BetaTableFPrimaryInformation{}$ & \BetaTableFEndpointAccuracy{} [\BetaTableFEndpointLower{}, \BetaTableFEndpointUpper{}] & \BetaTableFEndpointAUC{} & $\BetaTableFEndpointInformation{}$ \\
$2^{5}\ (\BetaTableGBeta{})$ & \BetaTableGPrimaryAccuracy{} [\BetaTableGPrimaryLower{}, \BetaTableGPrimaryUpper{}] & \BetaTableGPrimaryAUC{} & $\BetaTableGPrimaryInformation{}$ & \BetaTableGEndpointAccuracy{} [\BetaTableGEndpointLower{}, \BetaTableGEndpointUpper{}] & \BetaTableGEndpointAUC{} & $\BetaTableGEndpointInformation{}$ \\
\bottomrule
\end{tabular}

%% file: FiguresMake/generated/paired_table.tex
\begin{tabular}{@{}lrrrr@{}}
\toprule
Contrast and representation & $\Delta$ AUTOC & CI lower & CI upper & Holm $p$ \\
\midrule
Civic honesty / money vs.\ no money / wallet reported / area-pooled text & $\PairedTableTC{}$ & $\PairedTableTD{}$ & $\PairedTableTE{}$ & $\PairedTableTF{}$ \\
Civic honesty / money vs.\ no money / wallet reported / tabular + area text & $\PairedTableTG{}$ & $\PairedTableTH{}$ & $\PairedTableTI{}$ & $\PairedTableTJ{}$ \\
Direct air capture / no community benefits / profile selected / tabular + area text & $\PairedTableWI{}$ & $\PairedTableWJ{}$ & $\PairedTableWK{}$ & $\PairedTableWL{}$ \\
Direct air capture / enhanced oil recovery storage / profile selected / area-pooled text & $\PairedTableWU{}$ & $\PairedTableWV{}$ & $\PairedTableWW{}$ & $\PairedTableWX{}$ \\
Kenya digital ID / social protection / profile selected / text & $\PairedTableAGY{}$ & $\PairedTableAGZ{}$ & $\PairedTableAHA{}$ & $\PairedTableAHB{}$ \\
Kenya digital ID / social protection / profile selected / tabular + text & $\PairedTableAHC{}$ & $\PairedTableAHD{}$ & $\PairedTableAHE{}$ & $\PairedTableAHF{}$ \\
Kenya digital ID / tax registration / profile selected / tabular + text & $\PairedTableAHS{}$ & $\PairedTableAHT{}$ & $\PairedTableAHU{}$ & $\PairedTableAHV{}$ \\
Moral Machine / more vs.\ fewer characters / characters spared / text & $\PairedTableALW{}$ & $\PairedTableALX{}$ & $\PairedTableALY{}$ & $\PairedTableALZ{}$ \\
Moral Machine / more vs.\ fewer characters / characters spared / area-pooled text & $\PairedTableAME{}$ & $\PairedTableAMF{}$ & $\PairedTableAMG{}$ & $\PairedTableAMH{}$ \\
Trolley dilemmas / footbridge vs.\ switch / sacrifice acceptable / text & $\PairedTableANC{}$ & $\PairedTableAND{}$ & $\PairedTableANE{}$ & \scinum{\PairedTableANF{}} \\
Trolley dilemmas / footbridge vs.\ switch / sacrifice acceptable / area-pooled text & $\PairedTableANK{}$ & $\PairedTableANL{}$ & $\PairedTableANM{}$ & \scinum{\PairedTableANN{}} \\
Trolley dilemmas / footbridge vs.\ switch / sacrifice acceptable / tabular + area text & $\PairedTableANO{}$ & $\PairedTableANP{}$ & $\PairedTableANQ{}$ & $\PairedTableANR{}$ \\
Trolley dilemmas / loop vs.\ switch / sacrifice acceptable / text & $\PairedTableANS{}$ & $\PairedTableANT{}$ & $\PairedTableANU{}$ & $\PairedTableANV{}$ \\
Trolley dilemmas / loop vs.\ switch / sacrifice acceptable / area-pooled text & $\PairedTableAOA{}$ & $\PairedTableAOB{}$ & $\PairedTableAOC{}$ & $\PairedTableAOD{}$ \\
\bottomrule
\end{tabular}

%% file: FiguresMake/generated/contrast_table.tex
\begin{longtable}{>{\raggedright\arraybackslash}p{0.51\linewidth}rrrrr}
\caption{\textbf{Primary AUTOC/SE by treatment--outcome contrast.}}\label{tab:contrast-detail}\\
\toprule
 & & \multicolumn{2}{c}{Text} & \multicolumn{2}{c}{Tabular + text} \\
\cmidrule(lr){3-4}\cmidrule(lr){5-6}
Study / contrast / outcome & Tabular & \begin{tabular}[b]{@{}r@{}}Session-isolated\\retrieval\end{tabular} & \begin{tabular}[b]{@{}r@{}}Workspace\\teammate\end{tabular} & \begin{tabular}[b]{@{}r@{}}Session-isolated\\retrieval\end{tabular} & \begin{tabular}[b]{@{}r@{}}Workspace\\teammate\end{tabular} \\
\midrule
\endfirsthead
\multicolumn{6}{l}{\tablename~\thetable\ (continued)} \\
\toprule
 & & \multicolumn{2}{c}{Text} & \multicolumn{2}{c}{Tabular + text} \\
\cmidrule(lr){3-4}\cmidrule(lr){5-6}
Study / contrast / outcome & Tabular & \begin{tabular}[b]{@{}r@{}}Session-isolated\\retrieval\end{tabular} & \begin{tabular}[b]{@{}r@{}}Workspace\\teammate\end{tabular} & \begin{tabular}[b]{@{}r@{}}Session-isolated\\retrieval\end{tabular} & \begin{tabular}[b]{@{}r@{}}Workspace\\teammate\end{tabular} \\
\midrule
\endhead
Bangladesh masks / mask promotion / proper mask wearing & $\ContrastTableFL{}$ & $\ContrastTableFM{}$ & $\ContrastTableFN{}$ & $\ContrastTableFO{}$ & $\ContrastTableFP{}$ \\
Bangladesh masks / mask promotion / physical distancing & $\ContrastTableFQ{}$ & $\ContrastTableFR{}$ & $\ContrastTableFS{}$ & $\ContrastTableFT{}$ & $\ContrastTableFU{}$ \\
COVID-19 boosters / 90\% efficacy / profile accepted & $\ContrastTableFV{}$ & $\ContrastTableFW{}$ & $\ContrastTableFX{}$ & $\ContrastTableFY{}$ & $\ContrastTableFZ{}$ \\
COVID-19 boosters / two-year duration / profile accepted & $\ContrastTableGA{}$ & $\ContrastTableGB{}$ & $\ContrastTableGC{}$ & $\ContrastTableGD{}$ & $\ContrastTableGE{}$ \\
COVID-19 boosters / variant protection / profile accepted & $\ContrastTableGF{}$ & $\ContrastTableGG{}$ & $\ContrastTableGH{}$ & $\ContrastTableGI{}$ & $\ContrastTableGJ{}$ \\
COVID-19 boosters / mRNA manufacturer / profile accepted & $\ContrastTableGK{}$ & $\ContrastTableGL{}$ & $\ContrastTableGM{}$ & $\ContrastTableGN{}$ & $\ContrastTableGO{}$ \\
COVID-19 boosters / \$1,000 incentive / profile accepted & $\ContrastTableGP{}$ & $\ContrastTableGQ{}$ & $\ContrastTableGR{}$ & $\ContrastTableGS{}$ & $\ContrastTableGT{}$ \\
Civic honesty / money vs.\ no money / wallet reported & $\ContrastTableGU{}$ & $\ContrastTableGV{}$ & $\ContrastTableGW{}$ & $\ContrastTableGX{}$ & $\ContrastTableGY{}$ \\
Direct air capture / fossil-fuel owner / profile selected & $\ContrastTableGZ{}$ & $\ContrastTableHA{}$ & $\ContrastTableHB{}$ & $\ContrastTableHC{}$ & $\ContrastTableHD{}$ \\
Direct air capture / fossil-fuel energy / profile selected & $\ContrastTableHE{}$ & $\ContrastTableHF{}$ & $\ContrastTableHG{}$ & $\ContrastTableHH{}$ & $\ContrastTableHI{}$ \\
Direct air capture / no consultation / profile selected & $\ContrastTableHJ{}$ & $\ContrastTableHK{}$ & $\ContrastTableHL{}$ & $\ContrastTableHM{}$ & $\ContrastTableHN{}$ \\
Direct air capture / no local jobs / profile selected & $\ContrastTableHO{}$ & $\ContrastTableHP{}$ & $\ContrastTableHQ{}$ & $\ContrastTableHR{}$ & $\ContrastTableHS{}$ \\
Direct air capture / no community benefits / profile selected & $\ContrastTableHT{}$ & $\ContrastTableHU{}$ & $\ContrastTableHV{}$ & $\ContrastTableHW{}$ & $\ContrastTableHX{}$ \\
Direct air capture / enhanced oil recovery storage / profile selected & $\ContrastTableHY{}$ & $\ContrastTableHZ{}$ & $\ContrastTableIA{}$ & $\ContrastTableIB{}$ & $\ContrastTableIC{}$ \\
Direct air capture / private funding / profile selected & $\ContrastTableID{}$ & $\ContrastTableIE{}$ & $\ContrastTableIF{}$ & $\ContrastTableIG{}$ & $\ContrastTableIH{}$ \\
Holiday advertisements / Thanksgiving ads / post-Thanksgiving cases & $\ContrastTableII{}$ & $\ContrastTableIJ{}$ & $\ContrastTableIK{}$ & $\ContrastTableIL{}$ & $\ContrastTableIM{}$ \\
Holiday advertisements / Christmas ads / post-Christmas cases & $\ContrastTableIN{}$ & $\ContrastTableIO{}$ & $\ContrastTableIP{}$ & $\ContrastTableIQ{}$ & $\ContrastTableIR{}$ \\
IST-1 stroke / aspirin / dead or dependent (6 mo.) & $\ContrastTableIS{}$ & $\ContrastTableIT{}$ & $\ContrastTableIU{}$ & $\ContrastTableIV{}$ & $\ContrastTableIW{}$ \\
IST-1 stroke / aspirin / death (14 d) & $\ContrastTableIX{}$ & $\ContrastTableIY{}$ & $\ContrastTableIZ{}$ & $\ContrastTableJA{}$ & $\ContrastTableJB{}$ \\
IST-1 stroke / heparin / dead or dependent (6 mo.) & $\ContrastTableJC{}$ & $\ContrastTableJD{}$ & $\ContrastTableJE{}$ & $\ContrastTableJF{}$ & $\ContrastTableJG{}$ \\
IST-1 stroke / heparin / death (14 d) & $\ContrastTableJH{}$ & $\ContrastTableJI{}$ & $\ContrastTableJJ{}$ & $\ContrastTableJK{}$ & $\ContrastTableJL{}$ \\
Just transition / any income support / profile selected & $\ContrastTableJM{}$ & $\ContrastTableJN{}$ & $\ContrastTableJO{}$ & $\ContrastTableJP{}$ & $\ContrastTableJQ{}$ \\
Just transition / any benefit support / profile selected & $\ContrastTableJR{}$ & $\ContrastTableJS{}$ & $\ContrastTableJT{}$ & $\ContrastTableJU{}$ & $\ContrastTableJV{}$ \\
Just transition / relocation voucher / profile selected & $\ContrastTableJW{}$ & $\ContrastTableJX{}$ & $\ContrastTableJY{}$ & $\ContrastTableJZ{}$ & $\ContrastTableKA{}$ \\
Just transition / clean-energy retraining / profile selected & $\ContrastTableKB{}$ & $\ContrastTableKC{}$ & $\ContrastTableKD{}$ & $\ContrastTableKE{}$ & $\ContrastTableKF{}$ \\
Just transition / \$100k salary / profile selected & $\ContrastTableKG{}$ & $\ContrastTableKH{}$ & $\ContrastTableKI{}$ & $\ContrastTableKJ{}$ & $\ContrastTableKK{}$ \\
Just transition / short retraining time / profile selected & $\ContrastTableKL{}$ & $\ContrastTableKM{}$ & $\ContrastTableKN{}$ & $\ContrastTableKO{}$ & $\ContrastTableKP{}$ \\
Just transition / school \& community investment / profile selected & $\ContrastTableKQ{}$ & $\ContrastTableKR{}$ & $\ContrastTableKS{}$ & $\ContrastTableKT{}$ & $\ContrastTableKU{}$ \\
Kenya digital ID / public services / profile selected & $\ContrastTableKV{}$ & $\ContrastTableKW{}$ & $\ContrastTableKX{}$ & $\ContrastTableKY{}$ & $\ContrastTableKZ{}$ \\
Kenya digital ID / security / profile selected & $\ContrastTableLA{}$ & $\ContrastTableLB{}$ & $\ContrastTableLC{}$ & $\ContrastTableLD{}$ & $\ContrastTableLE{}$ \\
Kenya digital ID / social protection / profile selected & $\ContrastTableLF{}$ & $\ContrastTableLG{}$ & $\ContrastTableLH{}$ & $\ContrastTableLI{}$ & $\ContrastTableLJ{}$ \\
Kenya digital ID / tax registration / profile selected & $\ContrastTableLK{}$ & $\ContrastTableLL{}$ & $\ContrastTableLM{}$ & $\ContrastTableLN{}$ & $\ContrastTableLO{}$ \\
Kenya digital ID / voting / profile selected & $\ContrastTableLP{}$ & $\ContrastTableLQ{}$ & $\ContrastTableLR{}$ & $\ContrastTableLS{}$ & $\ContrastTableLT{}$ \\
Moral Machine / young vs.\ elderly / characters spared & $\ContrastTableLU{}$ & $\ContrastTableLV{}$ & $\ContrastTableLW{}$ & $\ContrastTableLX{}$ & $\ContrastTableLY{}$ \\
Moral Machine / female vs.\ male / characters spared & $\ContrastTableLZ{}$ & $\ContrastTableMA{}$ & $\ContrastTableMB{}$ & $\ContrastTableMC{}$ & $\ContrastTableMD{}$ \\
Moral Machine / fit vs.\ large / characters spared & $\ContrastTableME{}$ & $\ContrastTableMF{}$ & $\ContrastTableMG{}$ & $\ContrastTableMH{}$ & $\ContrastTableMI{}$ \\
Moral Machine / higher vs.\ lower status / characters spared & $\ContrastTableMJ{}$ & $\ContrastTableMK{}$ & $\ContrastTableML{}$ & $\ContrastTableMM{}$ & $\ContrastTableMN{}$ \\
Moral Machine / humans vs.\ pets / characters spared & $\ContrastTableMO{}$ & $\ContrastTableMP{}$ & $\ContrastTableMQ{}$ & $\ContrastTableMR{}$ & $\ContrastTableMS{}$ \\
Moral Machine / more vs.\ fewer characters / characters spared & $\ContrastTableMT{}$ & $\ContrastTableMU{}$ & $\ContrastTableMV{}$ & $\ContrastTableMW{}$ & $\ContrastTableMX{}$ \\
Moral Machine / lawful vs.\ unlawful / characters spared & $\ContrastTableMY{}$ & $\ContrastTableMZ{}$ & $\ContrastTableNA{}$ & $\ContrastTableNB{}$ & $\ContrastTableNC{}$ \\
Trolley dilemmas / footbridge vs.\ switch / sacrifice acceptable & $\ContrastTableND{}$ & $\ContrastTableNE{}$ & $\ContrastTableNF{}$ & $\ContrastTableNG{}$ & $\ContrastTableNH{}$ \\
Trolley dilemmas / loop vs.\ switch / sacrifice acceptable & $\ContrastTableNI{}$ & $\ContrastTableNJ{}$ & $\ContrastTableNK{}$ & $\ContrastTableNL{}$ & $\ContrastTableNM{}$ \\
Uganda YOP / cash grant (ITT) / training hours (log) & $\ContrastTableNN{}$ & $\ContrastTableNO{}$ & $\ContrastTableNP{}$ & $\ContrastTableNQ{}$ & $\ContrastTableNR{}$ \\
Uganda YOP / cash grant (ITT) / vocational training & $\ContrastTableNS{}$ & $\ContrastTableNT{}$ & $\ContrastTableNU{}$ & $\ContrastTableNV{}$ & $\ContrastTableNW{}$ \\
Uganda YOP / cash grant (ITT) / skilled trade & $\ContrastTableNX{}$ & $\ContrastTableNY{}$ & $\ContrastTableNZ{}$ & $\ContrastTableOA{}$ & $\ContrastTableOB{}$ \\
Uganda YOP / cash grant (ITT) / human capital & $\ContrastTableOC{}$ & $\ContrastTableOD{}$ & $\ContrastTableOE{}$ & $\ContrastTableOF{}$ & $\ContrastTableOG{}$ \\
Uganda YOP / cash grant (ITT) / income & $\ContrastTableOH{}$ & $\ContrastTableOI{}$ & $\ContrastTableOJ{}$ & $\ContrastTableOK{}$ & $\ContrastTableOL{}$ \\
\bottomrule
\end{longtable}

%% file: FiguresMake/generated/encoder_table.tex
\begin{longtable}{>{\raggedright\arraybackslash}p{0.49\linewidth}rrrrr}
\caption{\textbf{Uncompressed-embedding R-learner sensitivity across encoders.} Within-study mean AUTOC/SE against a shared tabular R-learner benchmark, with encoder dimensions and chunk pooling specified in Appendix~\ref{app:encoder}.}\label{tab:encoder-detail}\\
\toprule
Study / workflow / features & Tabular & Small, \HeadlinePCN{} PCs & Small (\SmallEmbeddingDim{}-d) & Large (\LargeEmbeddingDim{}-d) & MPNet (\MPNetEmbeddingDim{}-d) \\
\midrule
\endfirsthead
\multicolumn{6}{l}{\tablename~\thetable\ (continued)} \\
\toprule
Study / workflow / features & Tabular & Small, \HeadlinePCN{} PCs & Small (\SmallEmbeddingDim{}-d) & Large (\LargeEmbeddingDim{}-d) & MPNet (\MPNetEmbeddingDim{}-d) \\
\midrule
\endhead
Uganda YOP / session-isolated retrieval / text & $\EncoderTableAWH{}$ & $\EncoderTableAWI{}$ & $\EncoderTableAWJ{}$ & $\EncoderTableAWK{}$ & $\EncoderTableAWL{}$ \\
Uganda YOP / session-isolated retrieval / tabular + text & $\EncoderTableAWM{}$ & $\EncoderTableAWN{}$ & $\EncoderTableAWO{}$ & $\EncoderTableAWP{}$ & $\EncoderTableAWQ{}$ \\
Uganda YOP / workspace teammate / text & $\EncoderTableAWR{}$ & $\EncoderTableAWS{}$ & $\EncoderTableAWT{}$ & $\EncoderTableAWU{}$ & $\EncoderTableAWV{}$ \\
Uganda YOP / workspace teammate / tabular + text & $\EncoderTableAWW{}$ & $\EncoderTableAWX{}$ & $\EncoderTableAWY{}$ & $\EncoderTableAWZ{}$ & $\EncoderTableAXA{}$ \\
Kenya digital ID / session-isolated retrieval / text & $\EncoderTableAXB{}$ & $\EncoderTableAXC{}$ & $\EncoderTableAXD{}$ & $\EncoderTableAXE{}$ & $\EncoderTableAXF{}$ \\
Kenya digital ID / session-isolated retrieval / tabular + text & $\EncoderTableAXG{}$ & $\EncoderTableAXH{}$ & $\EncoderTableAXI{}$ & $\EncoderTableAXJ{}$ & $\EncoderTableAXK{}$ \\
Kenya digital ID / workspace teammate / text & $\EncoderTableAXL{}$ & $\EncoderTableAXM{}$ & $\EncoderTableAXN{}$ & $\EncoderTableAXO{}$ & $\EncoderTableAXP{}$ \\
Kenya digital ID / workspace teammate / tabular + text & $\EncoderTableAXQ{}$ & $\EncoderTableAXR{}$ & $\EncoderTableAXS{}$ & $\EncoderTableAXT{}$ & $\EncoderTableAXU{}$ \\
Direct air capture / session-isolated retrieval / text & $\EncoderTableAXV{}$ & $\EncoderTableAXW{}$ & $\EncoderTableAXX{}$ & $\EncoderTableAXY{}$ & $\EncoderTableAXZ{}$ \\
Direct air capture / session-isolated retrieval / tabular + text & $\EncoderTableAYA{}$ & $\EncoderTableAYB{}$ & $\EncoderTableAYC{}$ & $\EncoderTableAYD{}$ & $\EncoderTableAYE{}$ \\
Direct air capture / workspace teammate / text & $\EncoderTableAYF{}$ & $\EncoderTableAYG{}$ & $\EncoderTableAYH{}$ & $\EncoderTableAYI{}$ & $\EncoderTableAYJ{}$ \\
Direct air capture / workspace teammate / tabular + text & $\EncoderTableAYK{}$ & $\EncoderTableAYL{}$ & $\EncoderTableAYM{}$ & $\EncoderTableAYN{}$ & $\EncoderTableAYO{}$ \\
Just transition / session-isolated retrieval / text & $\EncoderTableAYP{}$ & $\EncoderTableAYQ{}$ & $\EncoderTableAYR{}$ & $\EncoderTableAYS{}$ & $\EncoderTableAYT{}$ \\
Just transition / session-isolated retrieval / tabular + text & $\EncoderTableAYU{}$ & $\EncoderTableAYV{}$ & $\EncoderTableAYW{}$ & $\EncoderTableAYX{}$ & $\EncoderTableAYY{}$ \\
Just transition / workspace teammate / text & $\EncoderTableAYZ{}$ & $\EncoderTableAZA{}$ & $\EncoderTableAZB{}$ & $\EncoderTableAZC{}$ & $\EncoderTableAZD{}$ \\
Just transition / workspace teammate / tabular + text & $\EncoderTableAZE{}$ & $\EncoderTableAZF{}$ & $\EncoderTableAZG{}$ & $\EncoderTableAZH{}$ & $\EncoderTableAZI{}$ \\
COVID-19 boosters / session-isolated retrieval / text & $\EncoderTableAZJ{}$ & $\EncoderTableAZK{}$ & $\EncoderTableAZL{}$ & $\EncoderTableAZM{}$ & $\EncoderTableAZN{}$ \\
COVID-19 boosters / session-isolated retrieval / tabular + text & $\EncoderTableAZO{}$ & $\EncoderTableAZP{}$ & $\EncoderTableAZQ{}$ & $\EncoderTableAZR{}$ & $\EncoderTableAZS{}$ \\
COVID-19 boosters / workspace teammate / text & $\EncoderTableAZT{}$ & $\EncoderTableAZU{}$ & $\EncoderTableAZV{}$ & $\EncoderTableAZW{}$ & $\EncoderTableAZX{}$ \\
COVID-19 boosters / workspace teammate / tabular + text & $\EncoderTableAZY{}$ & $\EncoderTableAZZ{}$ & $\EncoderTableBAA{}$ & $\EncoderTableBAB{}$ & $\EncoderTableBAC{}$ \\
Bangladesh masks / session-isolated retrieval / text & $\EncoderTableBAD{}$ & $\EncoderTableBAE{}$ & $\EncoderTableBAF{}$ & $\EncoderTableBAG{}$ & $\EncoderTableBAH{}$ \\
Bangladesh masks / session-isolated retrieval / tabular + text & $\EncoderTableBAI{}$ & $\EncoderTableBAJ{}$ & $\EncoderTableBAK{}$ & $\EncoderTableBAL{}$ & $\EncoderTableBAM{}$ \\
Bangladesh masks / workspace teammate / text & $\EncoderTableBAN{}$ & $\EncoderTableBAO{}$ & $\EncoderTableBAP{}$ & $\EncoderTableBAQ{}$ & $\EncoderTableBAR{}$ \\
Bangladesh masks / workspace teammate / tabular + text & $\EncoderTableBAS{}$ & $\EncoderTableBAT{}$ & $\EncoderTableBAU{}$ & $\EncoderTableBAV{}$ & $\EncoderTableBAW{}$ \\
Holiday advertisements / session-isolated retrieval / text & $\EncoderTableBAX{}$ & $\EncoderTableBAY{}$ & $\EncoderTableBAZ{}$ & $\EncoderTableBBA{}$ & $\EncoderTableBBB{}$ \\
Holiday advertisements / session-isolated retrieval / tabular + text & $\EncoderTableBBC{}$ & $\EncoderTableBBD{}$ & $\EncoderTableBBE{}$ & $\EncoderTableBBF{}$ & $\EncoderTableBBG{}$ \\
Holiday advertisements / workspace teammate / text & $\EncoderTableBBH{}$ & $\EncoderTableBBI{}$ & $\EncoderTableBBJ{}$ & $\EncoderTableBBK{}$ & $\EncoderTableBBL{}$ \\
Holiday advertisements / workspace teammate / tabular + text & $\EncoderTableBBM{}$ & $\EncoderTableBBN{}$ & $\EncoderTableBBO{}$ & $\EncoderTableBBP{}$ & $\EncoderTableBBQ{}$ \\
IST-1 stroke / session-isolated retrieval / text & $\EncoderTableBBR{}$ & $\EncoderTableBBS{}$ & $\EncoderTableBBT{}$ & $\EncoderTableBBU{}$ & $\EncoderTableBBV{}$ \\
IST-1 stroke / session-isolated retrieval / tabular + text & $\EncoderTableBBW{}$ & $\EncoderTableBBX{}$ & $\EncoderTableBBY{}$ & $\EncoderTableBBZ{}$ & $\EncoderTableBCA{}$ \\
IST-1 stroke / workspace teammate / text & $\EncoderTableBCB{}$ & $\EncoderTableBCC{}$ & $\EncoderTableBCD{}$ & $\EncoderTableBCE{}$ & $\EncoderTableBCF{}$ \\
IST-1 stroke / workspace teammate / tabular + text & $\EncoderTableBCG{}$ & $\EncoderTableBCH{}$ & $\EncoderTableBCI{}$ & $\EncoderTableBCJ{}$ & $\EncoderTableBCK{}$ \\
Moral Machine / session-isolated retrieval / text & $\EncoderTableBCL{}$ & $\EncoderTableBCM{}$ & $\EncoderTableBCN{}$ & $\EncoderTableBCO{}$ & $\EncoderTableBCP{}$ \\
Moral Machine / session-isolated retrieval / tabular + text & $\EncoderTableBCQ{}$ & $\EncoderTableBCR{}$ & $\EncoderTableBCS{}$ & $\EncoderTableBCT{}$ & $\EncoderTableBCU{}$ \\
Moral Machine / workspace teammate / text & $\EncoderTableBCV{}$ & $\EncoderTableBCW{}$ & $\EncoderTableBCX{}$ & $\EncoderTableBCY{}$ & $\EncoderTableBCZ{}$ \\
Moral Machine / workspace teammate / tabular + text & $\EncoderTableBDA{}$ & $\EncoderTableBDB{}$ & $\EncoderTableBDC{}$ & $\EncoderTableBDD{}$ & $\EncoderTableBDE{}$ \\
Trolley dilemmas / session-isolated retrieval / text & $\EncoderTableBDF{}$ & $\EncoderTableBDG{}$ & $\EncoderTableBDH{}$ & $\EncoderTableBDI{}$ & $\EncoderTableBDJ{}$ \\
Trolley dilemmas / session-isolated retrieval / tabular + text & $\EncoderTableBDK{}$ & $\EncoderTableBDL{}$ & $\EncoderTableBDM{}$ & $\EncoderTableBDN{}$ & $\EncoderTableBDO{}$ \\
Trolley dilemmas / workspace teammate / text & $\EncoderTableBDP{}$ & $\EncoderTableBDQ{}$ & $\EncoderTableBDR{}$ & $\EncoderTableBDS{}$ & $\EncoderTableBDT{}$ \\
Trolley dilemmas / workspace teammate / tabular + text & $\EncoderTableBDU{}$ & $\EncoderTableBDV{}$ & $\EncoderTableBDW{}$ & $\EncoderTableBDX{}$ & $\EncoderTableBDY{}$ \\
Civic honesty / session-isolated retrieval / text & $\EncoderTableBDZ{}$ & $\EncoderTableBEA{}$ & $\EncoderTableBEB{}$ & $\EncoderTableBEC{}$ & $\EncoderTableBED{}$ \\
Civic honesty / session-isolated retrieval / tabular + text & $\EncoderTableBEE{}$ & $\EncoderTableBEF{}$ & $\EncoderTableBEG{}$ & $\EncoderTableBEH{}$ & $\EncoderTableBEI{}$ \\
Civic honesty / workspace teammate / text & $\EncoderTableBEJ{}$ & $\EncoderTableBEK{}$ & $\EncoderTableBEL{}$ & $\EncoderTableBEM{}$ & $\EncoderTableBEN{}$ \\
Civic honesty / workspace teammate / tabular + text & $\EncoderTableBEO{}$ & $\EncoderTableBEP{}$ & $\EncoderTableBEQ{}$ & $\EncoderTableBER{}$ & $\EncoderTableBES{}$ \\
\bottomrule
\end{longtable}